\pdfoutput=1

\documentclass[11pt]{article}

\usepackage[final]{acl}
\usepackage{multirow}
\usepackage{multicol}
\usepackage{times}
\usepackage{latexsym,amssymb}

\usepackage[T1]{fontenc}

\usepackage[utf8]{inputenc}

\usepackage{microtype}

\usepackage{inconsolata}

\usepackage{graphicx}

\usepackage{xcolor,multirow}
\usepackage{xspace,amsmath}

\DeclareMathOperator*{\argmax}{argmax}

\newcommand{\mpc}{MPC\xspace}
\newcommand{\mpcSuff}{MPC++\xspace}
\newcommand{\ddc}{DD\xspace}
\newcommand{\dstc}{DU\xspace}
\newcommand{\precp}{\textit{P-Prec}\xspace}
\newcommand{\recallp}{\textit{P-Rec}\xspace}
 
\usepackage{listings}
\lstdefinestyle{prompt}{
  language=bash,
  basicstyle=\small,
  backgroundcolor=\color{gray!10},
  frame=none,
  numbers=none,
  columns=fullflexible,
  breaklines=true,
  breakatwhitespace=true,
  showstringspaces=false,
  xleftmargin=0pt,
  xrightmargin=0pt,
  aboveskip=10pt,
  belowskip=10pt,
  literate={~} {$\sim$}{1},
}

\title{Chat-Ghosting: A Comparative Study of Methods for Auto-Completion in Dialog Systems}

\author{\normalfont Sandeep Mishra\textsuperscript{1}, Anubhab Mandal\textsuperscript{1}, Bishal Santra\textsuperscript{2}, Tushar Abhishek\textsuperscript{2} \\
        \normalfont Pawan Goyal\textsuperscript{1}, Manish Gupta\textsuperscript{3} \\
        \textsuperscript{1}Indian Institute of Technology Kharagpur, India \\
        \textsuperscript{2}Microsoft Research, India\\
        \textsuperscript{3}Microsoft, India 
}
\begin{document}
\maketitle
\begin{abstract}

Ghosting, the ability to predict a user's intended text input for inline query auto-completion, is an invaluable feature for modern search engines and chat interfaces, greatly enhancing user experience. By suggesting completions to incomplete queries (or prefixes), ghosting aids users with slow typing speeds, disabilities, or limited language proficiency. Ghosting is a challenging problem and has become more important with the ubiquitousness of chat-based systems like ChatGPT, Copilot, etc. Despite the increasing prominence of chat-based systems utilizing ghosting, this challenging problem of Chat-Ghosting has received little attention from the NLP/ML research community. There is a lack of standardized benchmarks and relative performance analysis of deep learning and non-deep learning methods. We address this through an open and thorough study of this problem using four publicly available dialog datasets: two human-human (DailyDialog and DSTC7-Ubuntu) and two human-bot (Open Assistant and ShareGPT). We experiment with various existing query auto-completion methods (using tries), n-gram methods and deep learning methods, with and without dialog context. We also propose a novel entropy-based dynamic early stopping strategy. Our analysis finds that statistical n-gram models and tries outperform deep learning based models in terms of both model performance and inference efficiency for seen prefixes. For unseen queries, neural models like T5 and Phi-2 lead to better results. Adding conversational context leads to significant improvements in ghosting quality, especially for Open-Assistant and ShareGPT. We make code and data publicly available\footnote{\url{https://github.com/Anubhab2002/Chat-AutoSuggest-Project}\label{datacodefootnote}}. 
\end{abstract}

\section{Introduction}
\label{sec:intro}
\begin{figure}[!t]
    \centering
    \includegraphics[width=\columnwidth]{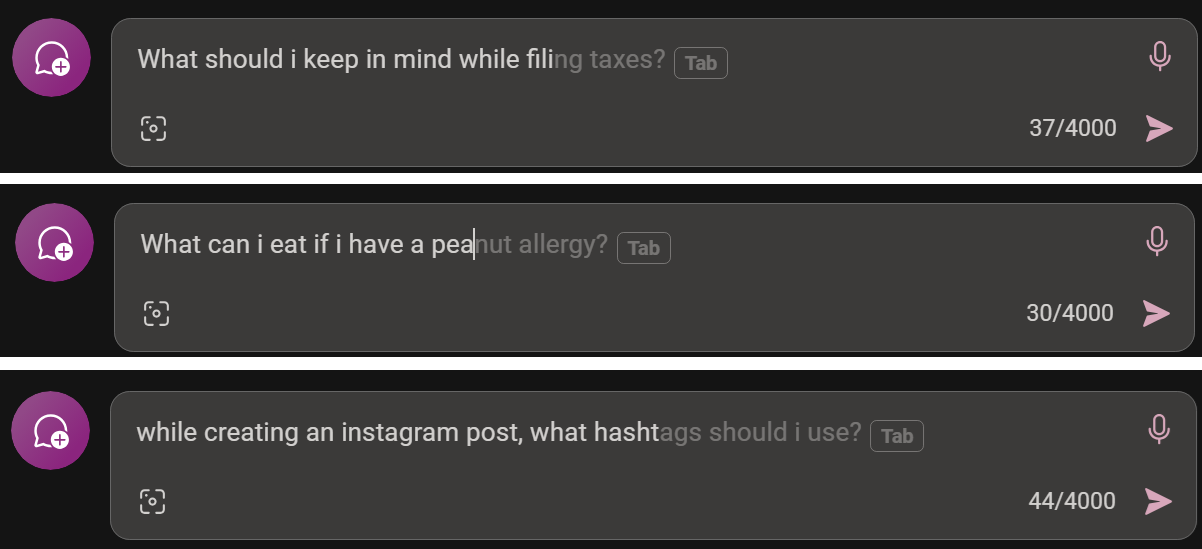}
    \caption{Ghosting Examples from Microsoft Copilot}
    \label{fig:example}
\end{figure}
AI-driven dialog systems or ``chatbots'' have become more important than ever before with the advent of LLM-based web interfaces like ChatGPT, Copilot, Gemini, etc. Here people can interact with models like GPT-4 and Gemini-Pro on both open-ended and domain-specific topics in the form of text-based chats. With the increasing popularity of such systems, it becomes imperative to make it easier for people to interact with these agents. 

One way to ease such interactions is via ghosting, the focus of this paper. Ghosting is a type-ahead completion task that predicts a user's intended input for inline query auto-completion (QAC). Given a partially typed query (the prefix), ghosting provides a single suggestion, unlike QAC, which typically shows $\sim$10. Fig.~\ref{fig:example} shows ghosting examples from Microsoft Copilot. Like QAC, ghosting helps users save time and, at times, offers creative suggestions that enable more expressive queries.

Ghosting is a challenging task. While being effective with short ambiguous prefixes, a ghosting system also needs to be efficient.
It should consider the context of the user input to generate relevant suggestions. Context could be noisy and tricky to incorporate for n-gram based models. Processing long contexts could also result into higher latency.

Although there is a vast literature on next utterance prediction using dialog models~\cite{zhang2019dialogpt,roller2020recipes,adiwardana2020towards}, the problem of completing the current utterance as the user types her prefix (i.e., ghosting), has been under-studied~\cite{ramachandran2019ghosting,chen2019gmail,trajanovski2021does,goren2024chai}. Hence, in this paper, we aim to properly define and establish a foundation for the chat-ghosting task. This includes setting up open standards for datasets and 
investigating effectiveness of existing language models for the task.

One of the major challenges for chat-ghosting is the lack of benchmark datasets. Hence, we adapt dialog response generation datasets for ghosting. Specifically, we curate two human-human conversation datasets, DailyDialog (DD)~\cite{li2017dailydialog} and DSTC7-Ubuntu (DU)~\cite{d2020overview}, to model both open-ended conversations and domain-specific conversations, respectively. Further, we leverage two human-bot datasets: Open Assistant (OASST)~\cite{kopf2024openassistant} and ShareGPT (SGPT)\footnote{\url{https://sharegpt.com/}, dataset version that was used: anon8231489123ShareGPT\_Vicuna\_unfiltered\label{sgptfn}} with long bot utterances.

Further, in this paper, we investigate several approaches to solve the chat-ghosting task, including QAC methods, n-gram models, neural language models, 
and prompt engineering. The most widely used QAC method is the most popular completion (MPC) supported by tries built on historical query logs. Query Blazer (QB)~\cite{kang2021queryblazer} is an n-gram method that consists of a subword unit encoder and a classical n-gram language model to generate query completions. In addition, we also fine-tune Transformer-based language models such as GPT2~\cite{brown2020language} and a sequence-to-sequence model such as T5~\cite{roberts2019exploring}. Lastly, we conduct experiments with prompting-based models such as Phi-2~\cite{gunasekar2023textbooks}, Mistral-7B~\cite{jiang2023mistral} and GPT4~\cite{achiam2023gpt}. 
We experiment with these models under the context-based setting (where we use previous few utterances in the conversation as supporting context) and without context. For QAC methods and n-gram-based models, we use context for reranking. But for neural language models and prompt engineering-based methods, we condition the generation on the context by prepending it to the prefix. 

Although ghosting is a natural language generation (NLG) task, standard NLG metrics like BLEU, ROUGE, CIDEr etc. are not suitable for this task as we need to rely on \textit{prefix-match} instead of n-gram or semantic matches. 
We measure output quality based on exact or partial match of the prediction with respect to the ground truth, using metrics like Match rate (MR), Partial Recall (\recallp), Partial Precision (\precp). Further, the tradeoff between the trigger rate (TR) and these quality metrics can vary significantly across methods. Hence, 
for fair comparison across methods, we capture quality metrics at \textit{maximum} TR for our main results. That said, we also show the tradeoff by varying TR for various methods. We measure the number of keystrokes the user is saved from typing using typing effort saved (TES) inspired by metrics defined in~\newcite{trajanovski2021does}. For completeness, we also present evaluation using METEOR and using GPT-4o judgments. 
Since low latency is crucial for ghosting, we also report the same for various methods.

We make the following main contributions: 
    (1) To the best of our knowledge, this is the first comprehensive work on ghosting for dialog/chat bot applications, providing a novel perspective on text prediction.
    (2) We investigate a diverse set of methodologies for this task both with and without context: trie-based most popular completion, n-gram-based QueryBlazer (QB), Transformer-based finetuned models like T5 and GPT2, and prompt-engineering methods like Phi-2, Mistral-7B and GPT4.
    (3) We introduce an entropy-based dynamic early stopping strategy that adaptively halts generation when model confidence drops, yielding more precise and efficient suggestions compared to static truncation.
    (4) Through a comprehensive comparative study on prominent datasets such as DD (open-domain), DU (task-specific), OASST and SGPT via multiple evaluation metrics, we compare the generation quality and latency of chat-ghosting across different methods, shedding light on their performance dynamics and computational efficiency.

\setlength{\tabcolsep}{4pt}
\begin{table*}[!t]
\scriptsize
\centering
\begin{tabular}{|l|c|c|c|c|c|c|c|c|}
\hline
& DD Train & DD Test & DU Train & DU Test & OASST Train & OASST Test & SGPT Train & SGPT Test\\ \hline
Avg. words per utterance & 12.37 & 12.42 & 13.48 & 14.55&20.36&23.35&53.27&56.85 \\ \hline
Avg. characters per utterance & 57.05 & 57.33 & 73.96 & 79.57 &115.05&131.85&341.01&366.88\\ \hline
No. of utterances & 69,216 & 7,986 & 549,002 & 5,588 &19421&981&328078&1088\\ \hline
\end{tabular}%
\caption{Statistics of the all 4 Datasets - DD, DU, OASST \& SGPT}
\label{tab:data_stats}
\end{table*}
\setlength{\tabcolsep}{6pt}

\section{Related Work}
\label{sec:relatedWork}
\noindent\textbf{Query Auto-Completion}: 
\citet{bast2006type} first introduced the efficacy of QAC in search engines. Ranking for QAC systems is typically supported by several modules such as most popular completion~\cite{whiting2013exploring}, time sensitive suggestions~\cite{shokouhi2012time,wang2017learning}, location sensitive suggestions~\cite{backstrom2008spatial,welch2008automatically}, ghosting~\cite{ramachandran2019ghosting}, session co-occurrences~\cite{bar2011context} and non-prefix matches~\cite{gog2020efficient}. 
Besides these, suggestion ranking in QAC has also been studied using traditional Machine Learning (ML) methods~\cite{di2015comparing,jiang2014learning,sordoni2015hierarchical} and deep learning (DL) methods like convolutional latent semantic model~\cite{mitra2015query}, LSTMs~\cite{wang2020efficient} and BERT and BART~\cite{mustar2020using}. Personalization has also been studied for QAC systems~\cite{shokouhi2013learning,song2017hierarchical,fiorini2018personalized,yin2020learning}. There have also been attempts to \textit{generate} effective QAC suggestions~\cite{park2017neural,kim2019subword,jiang2018rin,sordoni2015hierarchical,lee2021improving,olteanu2020search}. In this work, we focus on the chat-ghosting task. Unlike showing multiple full completions in the QAC systems, ghosting aims at generating a single relevant suggestion with next few  tokens.

\noindent\textbf{Dialog Response Generation}: Several Sequence-to-Sequence models~\citep{ritter2011data,santra2021hierarchical}, latent variable models~\cite{zhao2017learning,bao2019plato}, pretrained models~\cite{zhang2019dialogpt,roller2020recipes,adiwardana2020towards}, response retrieval/next-utterance selection models \cite{lowe2015ubuntu,chen2019sequential,whang2021response,xu2021learning,humeaupoly,henderson2020convert} and retrieval augmented generation~\cite{cai2021exemplar,komeili2021internet} models have been proposed for dialog response generation. In this work, we propose the task of ghosting for dialog systems. While dialog generation aims to generate full utterances, ghosting aims to generate next few tokens given a few characters already typed by the user. 

\noindent We discuss detailed related work in Appendix~\ref{app:detailedRelatedWork}.

\section{Chat-Ghosting: Task and Datasets}
\label{sec:method}
Ghosting aims to suggest accurate query completions 
based on an incomplete, partial prefix.

\noindent\textbf{Task Formulation}. 
The input to a chat-ghosting model comprises a partially written user prefix ($p$) and optional dialog history or context ($h$). 
The model is expected to predict a suitable completion ($c$) for the prefix $p$ such that $[p;c]$ is a valid response for dialog context $h$. 
Thus, at train time, the aim is to learn parameters $\theta$ of the chat-ghosting model as 
    $\theta^* = \argmax_\theta P(c|p,h,\theta)$. 
The trained model can then be used to generate predictions using $\hat{c} = \argmax_c P(c|p,h,\theta^*)$.

\noindent\textbf{Dataset Design}.
\label{sec:dataset}
There are no public datasets for training chat-ghosting models, complicating both training and evaluation. To address this, we simulate user interactions with auto-completion systems using existing dialog response generation datasets and emulate chat-ghosting outcomes.

Users typically engage in open-ended or domain-specific conversations with chatbots, making it necessary to train and evaluate the performance of the ghosting methods on both kinds of data. Hence, we use DailyDialog (DD), an open domain dialog dataset, and DSTC7-Ubuntu (DU), a domain-specific conversation dataset. 

DD contains open-domain English conversations, and we use a cleaned version to avoid data leaks \cite{wen2022empirical}. DU includes domain-specific English conversations from a Linux IRC channel. 

But DD and DU include human-human conversations. Hence, we also experiment with two other datasets: OASST and SGPT which include human-bot conversations. OASST is a human-annotated assistant conversation corpus. SGPT contains user-LLM-chatbots conversations collected by the ShareGPT API. Table~\ref{tab:data_stats} presents the basic statistics of these datasets. Since the main aim of ghosting is to assist a human, for both OASST and SGPT, we just consider just the human utterances for both training and inference to measure the effectiveness.

\section{Methods for Chat-Ghosting}
We follow various modeling strategies for chat-ghosting that gives a complete view at the spectrum of trade-off between accuracy, latency and coverage. We investigate several ML and DL approaches for the task, including QAC methods, n-gram models, neural language models and prompt engineering. We experiment with both the ``with context'' and ``without context'' settings. 

\noindent\textbf{Standard QAC Methods}. 
We experiment with two standard QAC methods: MPC and \mpcSuff.
Given a prefix, the MPC model, proposed in \newcite{bar2011context}, extracts a limited number ($k$) of completions from a character-trie structure (also referred to as the main trie) built using a corpus of past user-utterances. 
Prefixes for which completions are available through MPC are termed as `seen prefixes', while those lacking completions are termed as `unseen prefixes'. 

Main trie provides good coverage for QAC, however, for dialog systems, it is less likely that incoming user utterances would match with those in the training set. Hence, the coverage of the system can be very low due to novel utterances. 
To improve the coverage, we design an alternative method that utilizes a suffix-trie to handle unseen prefixes. The suffix trie is built using word n-gram suffixes (with freq $\geq$ 2) of the training utterances. 
Both the main trie and the suffix tries are character-level tries. The two tries can also be used in combination. Thus, in the \mpcSuff method, seen prefixes utilize the main trie, whereas unseen prefixes utilize the suffix trie. 

\noindent\textbf{N-Gram Model: Query Blazer (QB)}. 
QB is a fully-generative QAC system that uses a longest prefix match (LPM) subword unit encoder and a subword-level n-gram language model to generate query completions. QB can suggest queries from log history and generate new words or phrases. By leveraging the n-gram model’s data structure, QB materializes partial completions offline, reducing runtime overhead.

During training, QB extracts a subword vocabulary from the dataset and constructs an encoder as a finite state transducer (FST). The trained language model is represented as a weighted FST. 
During inference, based on the input prefix, the language model performs beam search to generate the final top-k completions. The confidence score for a prediction is the sum of the negative log-likelihoods of the predicted tokens, normalized by prediction length. This confidence score is used as a threshold for triggering decisions.

\noindent\textbf{Neural Language Modeling Methods}. We also experiment with popular Transformer-based models for generating ghosting suggestions. Specifically, we use the GPT2-base~\cite{brown2020language} (decoder-only) and T5-base ~\cite{roberts2019exploring} (Seq2Seq) models. GPT2-base has 12 decoder layers, while the T5-base has 12 encoder and 12 decoder layers. Our objective is to express the ghosting task as a causal language modeling task and investigate the efficacy of these models. 

The prefix and the completion constitute the source and target sequences, respectively, for these models.
The input prefix may end with a complete word, with a space after a complete word, or in the middle of an incomplete word. For example, for the query ``How are you'', consider the three cases where ``prefix$|$suffix'' pairs are (1) ``How are$|$ you'', (2) ``How are $|$you'', (3) ``How ar$|$e you''. In the first case, the model needs to learn to predict a space first while in the second and third cases, the model must generate a valid character. However, T5 and GPT2 tokenizers differ in the way they handle spaces, requiring us to handle the three cases carefully as discussed in Appendix~\ref{app:spaceHandling}.

\noindent\textbf{Prompt Engineering Methods}. 
We experiment with Large language models (LLMs) like GPT-4~\cite{achiam2023gpt}, and small language models (SLMs) like Phi-2~\cite{gunasekar2023textbooks} and Mistral-7B~\cite{jiang2023mistral} in zero-shot mode. We also finetune (FT) Phi-2 on our training datasets. Our experiments show that we could obtain best results by prompting the models to generate a max of 2 words for Mistral, 10 words for Phi-2 (PT) and 3 words for Phi-2 (FT).
In the zero-shot mode, we prompt (Appendix~\ref{app:nonContextualprompt}) the models to generate appropriate completions given a prefix. 

\noindent\textbf{Contextual Ghosting}. In chatbots, one can expect ghosting quality to significantly improve by considering previous utterances in the current conversation as additional context. Thus, we modify the methods described so far in this section to make use of previous user utterances while generating ghosting suggestions.

For neural language modeling methods and prompt engineering models, we prepend the context to the current user prefix. For prompt engineering-based methods, the detailed prompt is in Appendix~\ref{app:contextualprompt}. 
To incorporate historical context into QAC and n-gram models (QB, MPC, MPC++), we use a reranking setup. For each prefix in the test set, we first retrieve the top-$k$ completions $(k = 10)$ and their corresponding scores.
Next, we calculate the cosine similarity between each $(\text{prefix} + \text{completion})$ pair and the historical context. This cosine similarity is computed between the TF-IDF vectors of the text pairs, using a TF-IDF vectorizer trained on the training set to capture a dataset-specific vocabulary. Score for each completion is then calculated as $\alpha \cdot (\text{original\_score\_scaled}) + \beta \cdot (\text{cos\_sim}) \notag \\
   + \gamma \cdot (\text{length\_penalty\_scaled})$,
   where each term is scaled to $(-1, 1)$ for consistency across components. The hyperparameters $\alpha$, $\beta$, and $\gamma$ are tuned on a validation set, with $(\alpha, \beta, \gamma) = (0.5, 0.3, 0.2)$ yielding the best results.
Length penalty is calculated as $\frac{1}{1 + \text{completion\_length}}$. It helps to reduce the bias towards longer completions, which tend to have higher cosine similarity due to more matching words. The length penalty helps prioritize shorter, more relevant completions.

\setlength{\tabcolsep}{2pt}
\begin{table*}[!t]
\centering
\scriptsize
\begin{tabular}{|l|rrrrp{0.9cm}p{0.6cm}|rrrrp{0.9cm}p{0.6cm}|rrrrp{0.9cm}p{0.6cm}|}
\hline
\multicolumn{1}{|c|}{\multirow{2}{*}{\textbf{Models}}} & \multicolumn{6}{c|}{\textit{Unseen (94\%)}} & \multicolumn{6}{c|}{\textit{Seen (6\%)}} & \multicolumn{6}{c|}{\textit{Full}} \\ \cline{2-19} 
\multicolumn{1}{|c|}{}& \textit{MR} & \recallp & \precp & TES&Pred Len& MLen& \textit{MR} & \recallp & \precp   & TES&Pred Len& MLen& \textit{MR} & \recallp & \precp &TES& Pred Len& MLen \\ \hline
\textbf{QB}                      & 5.87          & 10.72          & \textbf{43.31} & \textbf{31.63} & 4.1           & 1.4          & 26.10          & 34.81          & 57.69          & 41.92          & 4.5           & 2.3          & 6.28          & 11.21          & \textbf{43.60} & \textbf{32.24} & 4.1           & 1.4          \\
\textbf{\mpc}     & 0.00          & 7.54           & 19.47          & 2.90           & \textbf{29.8} & 2.1          & \textbf{48.92} & \textbf{61.10} & \textbf{62.77} & \textbf{46.13} & 14.4          & \textbf{7.5}          & 4.30          & 11.98          & 23.06          & 5.50           & \textbf{28.5} & \textbf{2.6} \\
\textbf{\mpcSuff} & 5.14          & 11.77          & 28.49          & 13.85          & 19.2          & 1.8          & \textbf{48.92} & \textbf{61.10} & \textbf{62.77} & \textbf{46.13} & 14.4          & \textbf{7.5} & 6.41          & 13.20          & 29.49          & 15.79          & 19.1          & 2.0          \\
\textbf{T5 (220M)}               & \textbf{6.10} & 13.86          & 25.16          & 20.88          & 18.9          & 2.2          & 32.07          & 44.30          & 49.70          & 44.07          & 10.7          & 3.5          & \textbf{6.61} & 14.45          & 25.64          & 22.27          & 18.7          & 2.3          \\
\textbf{GPT2 (124M)}             & 5.26          & 11.66          & 27.64          & 24.47          & 12.3          & 1.9          & 28.50          & 39.14          & 50.61          & 45.30          & 8.0           & 2.9          & 5.72          & 12.20          & 28.10          & 25.72          & 12.3          & 1.9          \\
\textbf{Phi-2 (2.7B) (PT)}       & 2.82          & 6.00           & 17.21          & 12.38          & 10.7          & 0.9          & 15.01          & 25.24          & 34.72          & 22.84          & 10.6          & 1.8          & 3.10          & 6.44           & 17.61          & 13.01          & 10.7          & 0.9          \\
\textbf{Phi-2 (2.7B) (FT)}       & 0.90          & \textbf{15.59} & 27.54          & 13.12          & 9.0           & \textbf{2.5} & 3.52           & 38.59          & 33.12          & 15.07          & 8.5           & 2.8          & 0.95          & \textbf{16.04} & 27.65          & 13.23          & 9.0           & 2.5          \\
\textbf{Mistral-7B (PT)}         & 1.07          & 6.03           & 8.04           & 5.74           & 17.1          & 0.9          & 6.04           & 24.17          & 16.27          & 10.43          & \textbf{20.1} & 1.7          & 1.17          & 6.40           & 8.21           & 6.02           & 17.1          & 0.9          \\
\textbf{GPT4}                    & 3.16          & 13.81          & 18.52          & 12.11          & 18.0          & 2.3          & 4.98           & 33.00          & 20.77          & 10.82          & 18.8          & 2.6          & 3.20          & 14.19          & 18.56          & 12.03          & 18.0          & 2.3         
\\ \hline
\end{tabular}
\caption{Results of various approaches on the Chat-Ghosting task for the \ddc dataset. PT=Pretrained, FT=Finetuned. Metrics at max TR possible for each model as shown in Table~\ref{tab:coverage}. Pred Len and Matched Len (MLen) are in chars.}
\label{tab:results-ddc}
\end{table*}

\setlength{\tabcolsep}{2pt}
\begin{table*}[!t]
\centering
\scriptsize
\begin{tabular}{|l|rrrrp{0.9cm}p{0.6cm}|rrrrp{0.9cm}p{0.6cm}|rrrrp{0.9cm}p{0.6cm}|}
\hline
\multicolumn{1}{|c|}{\multirow{2}{*}{\textbf{Models}}} & \multicolumn{6}{c|}{\textit{Unseen (48.2\%)}} & \multicolumn{6}{c|}{\textit{Seen (51.8\%)}} & \multicolumn{6}{c|}{\textit{Full}} \\ \cline{2-19} 
\multicolumn{1}{|c|}{}& \textit{MR} & \recallp & \precp & TES&Pred Len& MLen& \textit{MR} & \recallp & \precp   & TES&Pred Len& MLen& \textit{MR} & \recallp & \precp &TES& Pred Len& MLen \\ \hline
\textbf{QB}                      & 2.31          & 8.05           & \textbf{39.77} & \textbf{24.70} & 5.3           & 1.7          & 6.22           & 12.26          & 47.46          & 32.43          & 6.1           & 2.7           & 4.29           & 10.19          & 43.66          & 28.70          & 5.7           & 2.2           \\
\textbf{\mpc}     & 0.00          & 5.73           & 15.34          & 2.01           & \textbf{53.8} & 2.0          & \textbf{86.49} & \textbf{87.37} & \textbf{88.81} & \textbf{74.07} & \textbf{71.0} & \textbf{64.5} & \textbf{75.07} & \textbf{76.57} & \textbf{79.09} & 39.24          & \textbf{68.7} & \textbf{56.2} \\
\textbf{\mpcSuff} & 2.03          & 8.44           & 26.82          & 10.42          & 32.4          & 1.9          & \textbf{86.49} & \textbf{87.37} & \textbf{88.81} & \textbf{74.07} & \textbf{71.0} & \textbf{64.5} & 54.22          & 57.21          & 65.13          & \textbf{43.30} & 56.2          & 40.6          \\
\textbf{T5 (220M)}               & \textbf{3.93} & \textbf{11.76} & 24.59          & 16.82          & 26.1          & \textbf{2.9} & 5.86           & 13.51          & 26.07          & 21.61          & 27.5          & 3.4           & 4.90           & 12.64          & 25.34          & 19.30          & 26.8          & 3.2           \\
\textbf{GPT2 (124M)}             & 3.18          & 9.98           & 25.08          & 19.02          & 20.3          & 2.4          & 7.84           & 14.54          & 30.06          & 27.26          & 22.3          & 4.1           & 5.53           & 12.28          & 27.60          & 23.28          & 21.3          & 3.3           \\
\textbf{Phi-2 (2.7B) (PT)}       & 0.76          & 4.59           & 19.37          & 9.81           & 13.1          & 1.1          & 1.00           & 5.06           & 20.42          & 10.73          & 13.3          & 1.2           & 0.88           & 4.83           & 19.90          & 10.29          & 13.2          & 1.2           \\
\textbf{Phi-2 (2.7B) (FT)}       & 0.67          & 11.08          & 31.92          & 13.37          & 8.3           & 2.5          & 0.80           & 11.38          & 32.11          & 13.74          & 8.2           & 2.5           & 0.74           & 11.23          & 32.02          & 13.56          & 8.3           & 2.5           \\
\textbf{Mistral-7B (PT)}         & 0.25          & 3.78           & 7.51           & 3.90           & 18.5          & 1.0          & 0.28           & 3.99           & 7.63           & 3.75           & 18.4          & 1.0           & 0.26           & 3.88           & 7.57           & 3.82           & 18.5          & 1.0           \\
\textbf{GPT4}                    & 1.40          & 9.21           & 16.55          & 9.55           & 19.7          & 2.2          & 1.37           & 9.29           & 16.55          & 9.08           & 19.9          & 2.3           & 1.38           & 9.25           & 16.55          & 9.31           & 19.8          & 2.3          
\\ \hline
\end{tabular}
\caption{Results of various approaches on the Chat-Ghosting task for the \dstc dataset. PT=Pretrained, FT=Finetuned. Metrics at max TR possible for each model as shown in Table~\ref{tab:coverage}. Pred Len and Matched Len (MLen) are in chars.}
\label{tab:results-dstc7}
\end{table*}

\noindent\textbf{Ghosting Evaluation Metrics.}
\label{subsec:metrics}
Standard NLG metrics like BLEU, ROUGE, etc. are not suitable for ghosting as we need to rely on prefix-match instead of n-gram or semantic matches. 
Any deviation from prefix matches typically leads to higher cognitive load in examining the suggestion, leading to less likelihood of suggestion acceptance. Hence, we design new metrics suitable for ghosting.\\
\underline{Quality Metrics:} When a ghosting suggestion is shown, the suggestion text may match exactly with the ground truth or match just an initial part of the target. Hence, we define Match Rate (MR) as the ratio of number of times the suggestion exactly matched the ground truth to the number of times suggestions were shown, similar to~\newcite{trajanovski2021does}. For real world usability, MR is too strict as users may accept partially correct suggestions and make the necessary minor adjustments. Further, users may show some bias towards correctness on the prefix side of the prediction than near the end. Hence, to allow for partial matches, we define Partial Recall (\recallp) and Partial Precision (\precp). \precp (or \recallp) is defined as the fraction of the total character length of the prediction (or ground truth) by which the prefix of prediction and ground truth overlap. For completeness, we also present evaluation using METEOR and using GPT-4o judgments in Appendix~\ref{app:semantics}.\\
\underline{Trigger Rate (TR):} At every character, the ghosting system can show a suggestion or not based on the confidence score of the suggestion. For users with high typing speed, intent clarity and good language skills, frequent ghosting may be distracting. On the other hand, for users who are slow at typing or who struggle in query formulation, triggering ghosting at every character may be acceptable. 
Although defining a threshold based on user traits is out of scope of the current paper, we introduce TR as an operating point. It is the ratio of number of times a ghosting suggestion was shown to the number of characters typed by the user.\\
If TR<100\% (or whatever is max possible for a model), the probability of low confidence predictions will decrease and the system will suggest completions only for some subset of all the test cases.
Hence, for fair comparison of various modeling approaches, we compare their performances with respect to complete match (MR) or partial match (\recallp and \precp) metrics at max TR. \\
\underline{Typing Effort Saved (TES):} Given an utterance in the test set, we first list down all possible $\langle$prefix, completion$\rangle$ splits. Thus, an utterance of length $l$ leads to $l-1$ $\langle$prefix, completion$\rangle$ samples. Metrics like MR, \recallp and \precp are computed per prefix and micro-averaged across the entire test set. On the other hand, the TES metric is computed per utterance in a manner such that it simulates the interaction of the user with the auto-completion system, based on a greedy selection heuristic, until the entire utterance is typed out. At any character position in the utterance, the auto-completion system generates a completion and we assume that the user either (a) accepts it if the full suggestion of length $s$ matches exactly with the first $s$ characters of the target completion, or (b) rejects the suggestion if it does not match, and types the next character. This process continues until the full ground truth utterance has been generated. At the end of the process, the TES metric is computed as 1 minus the ratio of the number of characters typed by the user to the utterance length. For example, let the utterance be ``who am I?''. Let the process be as follows: user types ``w'', user accepts ``ho'', user types ``[space]'', user does not accept ``is'' and continues to type ``a'', user accepts ``m i?'' Here the user typed 3 characters and utterance length is 9. Hence, TES=1-(3/9)=2/3. 
Note that TES is closely related to Est. ChS~\cite{trajanovski2021does} and eSaved~\cite{kharitonov2013user}.

\section{Experiments and Results}
\label{sec:experiments}

We aim to determine which model performs the best for the chat-ghosting task in with/without and human-human vs human-bot settings. Additionally, we explore the trade-offs between model performance and coverage. Detailed hyper-parameter settings can be found in Appendix~\ref{app:hyperparams}. 

\begin{table}[!t]
    \centering
    \scriptsize
    \begin{tabular}{|l|c|c|c|c|c|c|}
    \hline
&\multicolumn{3}{c|}{DD}&\multicolumn{3}{c|}{DU}\\
\hline
&Unseen&Seen&Full&Unseen&Seen&Full\\
\hline
 \textbf{QB} &86.28&90.61&86.37&85.81&86.60&86.21\\
 \textbf{\mpc} &21.92&100.00&23.44&15.51&100.00&58.11\\
 \textbf{\mpcSuff} &66.25&100.00&66.91&62.88&100.00&81.60\\
 \textbf{T5 (220M)} &100.00&100.00&100.00&99.98&99.99&99.98\\
 \textbf{GPT2 (124M)} &95.41&97.57&95.45&96.46&96.87&96.67\\
 \textbf{Phi-2 (2.7B) (PT)} &59.14&70.08&59.35&49.73&48.92&49.32\\
 \textbf{Phi-2 (2.7B) (FT)} &99.49&99.83&99.50&90.33&91.69&91.02\\
 \textbf{Mistral-7B (PT)} &65.58&68.64&65.64&55.54&54.66&55.10\\
 \textbf{GPT4} &98.31&99.57&98.34&90.24&91.58&90.92\\
 \hline
    \end{tabular}
    \caption{Max TR values. Values<100\% imply that models generate empty predictions.}
    \label{tab:coverage}
\end{table}

\setlength{\tabcolsep}{1pt}
\begin{table*}[!t]
\scriptsize
\centering
\begin{tabular}{|l|llllllllllll|llllllllllll|}
\hline
 & \multicolumn{12}{c|}{DD} & \multicolumn{12}{c|}{DU} \\ \hline
 & \multicolumn{3}{c|}{Len=1-5 chars} & \multicolumn{3}{c|}{Len=6-12 chars} & \multicolumn{3}{c|}{Len=13-25 chars} & \multicolumn{3}{c|}{Len=26-50 chars} & \multicolumn{3}{c|}{Len=1-5 chars} & \multicolumn{3}{c|}{Len=6-12 chars} & \multicolumn{3}{c|}{Len=13-25 chars} & \multicolumn{3}{c|}{Len=26-50 chars} \\ \hline
 & \multicolumn{1}{l|}{MR} & \multicolumn{1}{l|}{P-Rec} & \multicolumn{1}{l|}{P-Prec} & \multicolumn{1}{l|}{MR} & \multicolumn{1}{l|}{P-Rec} & \multicolumn{1}{l|}{P-Prec} & \multicolumn{1}{l|}{MR} & \multicolumn{1}{l|}{P-Rec} & \multicolumn{1}{l|}{P-Prec} & \multicolumn{1}{l|}{MR} & \multicolumn{1}{l|}{P-Rec} & P-Prec & \multicolumn{1}{l|}{MR} & \multicolumn{1}{l|}{P-Rec} & \multicolumn{1}{l|}{P-Prec} & \multicolumn{1}{l|}{MR} & \multicolumn{1}{l|}{P-Rec} & \multicolumn{1}{l|}{P-Prec} & \multicolumn{1}{l|}{MR} & \multicolumn{1}{l|}{P-Rec} & \multicolumn{1}{l|}{P-Prec} & \multicolumn{1}{l|}{MR} & \multicolumn{1}{l|}{P-Rec} & P-Prec \\ \hline
\mpc & \multicolumn{1}{l|}{0.01} & \multicolumn{1}{l|}{4.61} & \multicolumn{1}{l|}{22.93} & \multicolumn{1}{l|}{0.06} & \multicolumn{1}{l|}{7.33} & \multicolumn{1}{l|}{17.82} & \multicolumn{1}{l|}{0.65} & \multicolumn{1}{l|}{11.71} & \multicolumn{1}{l|}{15.82} & \multicolumn{1}{l|}{8.17} & \multicolumn{1}{l|}{\textbf{29.56}} & 29.78 & \multicolumn{1}{l|}{0.04} & \multicolumn{1}{l|}{3.10} & \multicolumn{1}{l|}{20.00} & \multicolumn{1}{l|}{0.19} & \multicolumn{1}{l|}{6.14} & \multicolumn{1}{l|}{12.71} & \multicolumn{1}{l|}{0.33} & \multicolumn{1}{l|}{8.46} & \multicolumn{1}{l|}{11.06} & \multicolumn{1}{l|}{0.79} & \multicolumn{1}{l|}{\textbf{14.75}} & 17.81 \\ \hline
\mpcSuff & \multicolumn{1}{l|}{0.01} & \multicolumn{1}{l|}{4.61} & \multicolumn{1}{l|}{22.93} & \multicolumn{1}{l|}{0.39} & \multicolumn{1}{l|}{7.48} & \multicolumn{1}{l|}{18.86} & \multicolumn{1}{l|}{4.05} & \multicolumn{1}{l|}{12.08} & \multicolumn{1}{l|}{26.96} & \multicolumn{1}{l|}{\textbf{9.06}} & \multicolumn{1}{l|}{15.91} & 34.68 & \multicolumn{1}{l|}{0.04} & \multicolumn{1}{l|}{3.10} & \multicolumn{1}{l|}{20.00} & \multicolumn{1}{l|}{0.25} & \multicolumn{1}{l|}{6.08} & \multicolumn{1}{l|}{14.15} & \multicolumn{1}{l|}{1.59} & \multicolumn{1}{l|}{8.06} & \multicolumn{1}{l|}{23.28} & \multicolumn{1}{l|}{2.44} & \multicolumn{1}{l|}{9.07} & 30.92 \\ \hline
QB & \multicolumn{1}{l|}{\textbf{0.13}} & \multicolumn{1}{l|}{3.48} & \multicolumn{1}{l|}{\textbf{36.09}} & \multicolumn{1}{l|}{1.14} & \multicolumn{1}{l|}{5.49} & \multicolumn{1}{l|}{\textbf{42.38}} & \multicolumn{1}{l|}{4.80} & \multicolumn{1}{l|}{9.88} & \multicolumn{1}{l|}{\textbf{44.38}} & \multicolumn{1}{l|}{8.30} & \multicolumn{1}{l|}{13.52} & \textbf{44.80} & \multicolumn{1}{l|}{0.13} & \multicolumn{1}{l|}{3.29} & \multicolumn{1}{l|}{\textbf{29.05}} & \multicolumn{1}{l|}{0.95} & \multicolumn{1}{l|}{6.24} & \multicolumn{1}{l|}{\textbf{37.69}} & \multicolumn{1}{l|}{2.07} & \multicolumn{1}{l|}{8.01} & \multicolumn{1}{l|}{\textbf{43.38}} & \multicolumn{1}{l|}{2.77} & \multicolumn{1}{l|}{8.83} & \textbf{42.46} \\ \hline
T5& \multicolumn{1}{l|}{0.08} & \multicolumn{1}{l|}{\textbf{5.11}} & \multicolumn{1}{l|}{22.36} & \multicolumn{1}{l|}{\textbf{1.21}} & \multicolumn{1}{l|}{\textbf{8.34}} & \multicolumn{1}{l|}{21.97} & \multicolumn{1}{l|}{\textbf{5.31}} & \multicolumn{1}{l|}{\textbf{13.40}} & \multicolumn{1}{l|}{24.76} & \multicolumn{1}{l|}{8.55} & \multicolumn{1}{l|}{16.68} & 27.65 & \multicolumn{1}{l|}{\textbf{0.14}} & \multicolumn{1}{l|}{\textbf{3.83}} & \multicolumn{1}{l|}{20.67} & \multicolumn{1}{l|}{\textbf{0.96}} & \multicolumn{1}{l|}{\textbf{7.00}} & \multicolumn{1}{l|}{22.22} & \multicolumn{1}{l|}{\textbf{2.83}} & \multicolumn{1}{l|}{\textbf{9.99}} & \multicolumn{1}{l|}{24.90} & \multicolumn{1}{l|}{\textbf{4.16}} & \multicolumn{1}{l|}{11.49} & 25.32 \\ \hline
\end{tabular}
\caption{Results of various approaches on the Chat-Ghosting task, on the \ddc and \dstc dataset for different prefix lengths. Metrics at max TR and reported for unseen test set.}
\label{tab:results-ddc-dstc7-prefix-length}
\end{table*}

\setlength{\tabcolsep}{2pt}
\begin{table*}[!t]
    \centering
    \scriptsize
    \begin{tabular}{|l|l|p{0.15\textwidth}|p{0.23\textwidth}|p{0.2\textwidth}|}
    \hline
&\multicolumn{2}{c|}{DD}&\multicolumn{2}{c|}{DU}\\
\hline
&Seen&Unseen&Seen&Unseen\\
\hline
\textbf{Prefix}&listen to the &which kind of salad are you goi&unable to fetch some archives, maybe run apt-get update or&so do you know h\\
\hline
\textbf{Ground Truth (GT)}&sound of nature ! it's like music&ng to make&try with --fix-missing?&owto setup a static route in the routing table\\ 
\hline
 \textbf{QB} &radio&ng&try with --fix-missing?&ow?\\
 \textbf{\mpc} &sound of nature ! it's like music&[EMPTY]&try with --fix-missing?&ow?\\
 \textbf{\mpcSuff} &sound of nature ! it's like music&ng&try with --fix-missing?&ow?\\
 \textbf{T5 (220M)} &music, please&ng to make&something?&ow to do that?\\
 \textbf{GPT2 (124M)} &news&ng to make&acle-java7-jre will help&ow to do that?\\
 \textbf{Phi-2 (2.7B) (PT)} &music&ng to make&check for updates&ow to use a computer\\
 \textbf{Phi-2 (2.7B) (FT)} &music. it&ng to have&something?&ow to do\\
 \textbf{Mistral-7B (PT)} &radio station for classic hits&ng to make?&try using a different package manager&ow to make a paper airplane?\\
 \textbf{GPT4} &latest podcast&ng to make?&try a different mirror&ow to fix this problem?\\
 \hline
    \end{tabular}
    \caption{Examples of suffixes predicted by various non-contextual models for different prefixes. More examples from both contextual and non-contextual models are in Appendix~\ref{app:context-results}.}
    \label{tab:caseStudies}
\end{table*}

\noindent\textbf{Main Ghosting Results (without context)}. Tables~\ref{tab:results-ddc} and~\ref{tab:results-dstc7} show accuracy metrics and prediction lengths for unseen/seen/full test sets across various models on DD and DU, resp. Due to lack of space, we report results for OASST and SGPT in Tables~\ref{tab:results-oasst} and~\ref{tab:results-sgp} resp. in Appendix~\ref{app:oasstSGPResults}. Here ``seen/unseen'' implies that the utterance was seen/unseen in the train set. These results have been computed at max TR, i.e., no thresholding of confidence values was done to compute these results, and thus comparison is made with respect to all samples in the test sets. However, as shown in Table~\ref{tab:coverage}, the max TR may not be 100\% for all models as models may generate empty predictions for some queries. 

From Tables~\ref{tab:results-ddc} and~\ref{tab:results-dstc7}, several key observations emerge. For unseen queries, the QB model proves to be the most effective for ghosting, achieving the highest TES and \precp{} scores, followed by the MPC++, finetuned T5, and GPT-2 models. 
Specifically, QB has the highest \precp ($\sim$43 in \ddc and $\sim$40 in \dstc), indicating that on average, the first 42\% characters of its predictions are correct. However, QB’s \recallp remains low. 
Despite QB’s higher \precp{} compared to T5 or GPT-2, its individual completions are much shorter, averaging 4.1 and 5.3 characters for DD and DU, respectively. While T5 and GPT-2 may have longer prefix matches in their suggested completions, longer incorrect completions could negatively impact user experience.

\begin{figure*}[!t] 
\centering 
\begin{minipage}{\textwidth}
\centering
\includegraphics[width=0.49\textwidth]{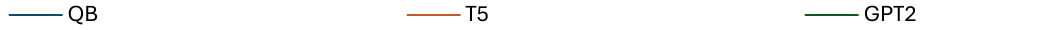} 
\includegraphics[width=0.49\textwidth]{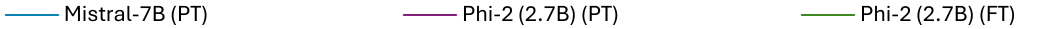} 
\end{minipage}
\begin{minipage}{0.24\textwidth}
\includegraphics[width=\columnwidth]{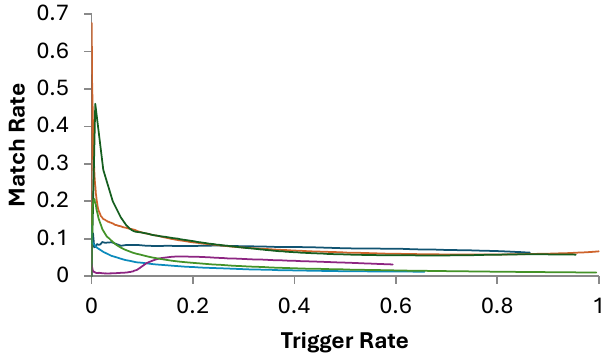} \caption{MR for DD} \label{fig:DD-MR}
\end{minipage}
\hfill
\begin{minipage}{0.24\textwidth}
\includegraphics[width=\columnwidth]{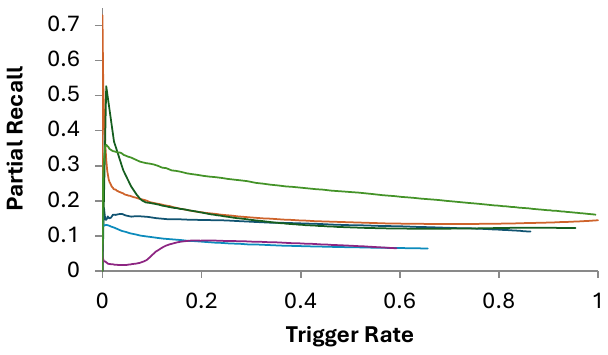} \caption{\recallp for DD} \label{fig:DD-P-Rec}
\end{minipage}
\hfill
\begin{minipage}{0.24\textwidth}
\includegraphics[width=\columnwidth]{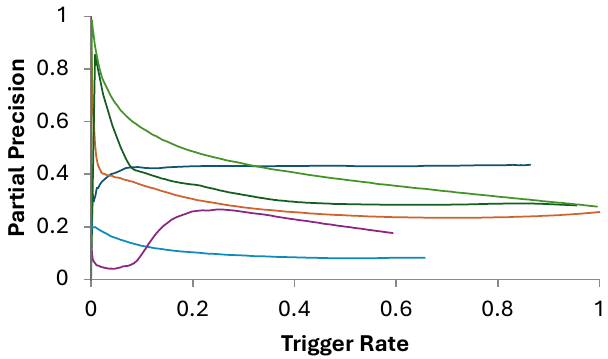} \caption{\precp for DD} \label{fig:DD-P-Prec}
\end{minipage}
\hfill
\begin{minipage}{0.24\textwidth}
\includegraphics[width=\columnwidth]{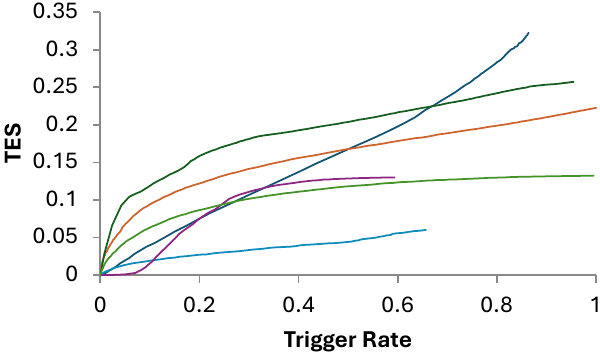} \caption{TES for DD} \label{fig:DD-TES} 
\end{minipage}
 \end{figure*}

When comparing ``trained'' language models (LMs) to other solutions like trie and prompt engineering, ``trained'' LMs emerge as the clear winners. Within this family, the classical frequency-based LM QB, likely due to its implicit memorization capabilities, outperforms the trained neural LMs such as GPT-2, T5, and Phi-2 (FT). As expected, trie-based solutions, which rely solely on memory, significantly outperform other models in the seen test set. MPC++, which combines trie with some language modeling capabilities (through the use of suffix-trie), performs better than MPC for unseen queries but falls short for seen queries. MPC also provides the longest suggestions. Considering both MR and TR, T5 is the best model for unseen queries, while finetuned Phi-2 outperforms pretrained Phi-2 except on MR.

Overall, these results suggest that for effective ghosting, it is beneficial to use LMs with inherent memory capacity and train them on logs. If a query seen during training is encountered during testing, the model reiterating the exact same completion is preferred over a model hallucination. Models that cannot be trained, such as zero-shot LLMs, are not effective for ghosting. It is important to note that all models perform poorly on the unseen test set, highlighting the challenging nature of chat-ghosting. Transformer-based LMs, regardless of their scale, also fail to perform satisfactorily. Zero-shot, instruction-tuned SLMs and LLMs perform significantly worse than their finetuned counterparts, underscoring the necessity of training with available logs for reliable performance. Based on our findings, we recommend using MPC for seen prefixes, and T5 or QB can be employed for unseen prefixes, depending on the latency requirements.

For OASST and SGPT, where utterance lengths are much longer, we observe that the pretrained knowledge helps and Mistral and Phi-2 provide the best \recallp and MR resp for the unseen set. Phi-2 also provides the best TES on unseen set. On the seen sets, \mpc, \mpcSuff and QB continue to perform better even on these human-bot datasets.

\noindent\textbf{Truncation-based results.} Various methods generate predictions of different sizes. Intuitively, longer suggestions have a lesser probability of being correct. This implies that MR and TES should be higher for shorter suggestions. To verify this we truncate the predicted suggestions to $t$ words where we vary $t$ from 1 to 10. We show results for MR, \precp, \recallp{} and TES with DD and DU datasets in contextual and non-contextual settings across all the methods in Figs.~\ref{fig:trunc-MR-DD} to~\ref{fig:trunc-TES-DU} in Appendix~\ref{app:truncationResults} for the unseen test set. GPT2 and T5 provide high TES (better than QB), when truncating to 1-2 words.

\noindent\textbf{Dynamic early stopping.} Static truncation of outputs to a fixed number of tokens may not be optimal across all test samples. 
A more adaptive approach is to halt generation when model confidence in the next token drops. This confidence can be estimated using entropy of the token probability distribution.

We apply entropy-based dynamic early stopping with thresholds of 0.6 and 3 during suggestion generation using T5 and GPT2 on the unseen test sets of DD and DU in the non-contextual setting. As shown in Table~\ref{tab:dynamicTruncation}, a threshold of 0.6 significantly boosts \precp{}, with T5 reaching 43.95 and GPT2 46.16 on the DD dataset. This improvement comes with shorter predictions, indicating more concise outputs, and a marked increase in TES, reflecting reduced keystrokes. Overall, dynamic early stopping offers a better trade-off across metrics compared to static truncation.

\begin{table}[!t]
    \centering
    \scriptsize
    \begin{tabular}{|l|l|c|c|c|c|c|c|c|}
    \hline
Model&Dataset&Threshold&TR&MR&P-Prec&P-Rec&Pred Len&TES\\
\hline
T5&DD&None&100.00&6.10&25.16&13.86&18.9&20.88\\
T5&DD&3&96.42&5.80&33.14&12.99&14.5&29.77\\
T5&DD&0.6&88.03&5.56&43.95&12.50&7.0&45.80\\
\cline{2-9}
T5&DU&None&99.98&3.93&24.59&11.76&26.1&16.82\\
T5&DU&3&95.92&3.71&33.81&11.06&21.0&27.99\\
T5&DU&0.6&91.62&2.94&48.97&9.19&11.6&44.41\\
\hline
GPT2&DD&None&95.41&5.26&27.64&11.66&12.3&24.47\\
GPT2&DD&3&97.56&4.61&35.83&10.41&9.2&34.75\\
GPT2&DD&0.6&99.38&3.81&46.16&8.71&4.1&45.83\\
\cline{2-9}
GPT2&DU&None&96.46&3.18&25.08&9.98&20.3&19.02\\
GPT2&DU&3&97.71&2.73&34.29&8.95&17.8&31.44\\
GPT2&DU&0.6&98.82&2.02&47.01&7.25&7.3&45.72\\
\hline
    \end{tabular}
    \caption{Entropy-based Dynamic Early Stopping}
    \label{tab:dynamicTruncation}
\end{table}

\noindent\textbf{Contextual Ghosting Results}. Tables~\ref{tab:results-ddc-contextual} and~\ref{tab:results-dstc7-contextual} in Appendix~\ref{app:context-results} show accuracy metrics and prediction lengths for unseen/seen/full test sets across various models on OASST and SGPT datasets, respectively for the ``with context'' setting. Table~\ref{tab:coverage-contextual} in Appendix~\ref{app:context-results} shows max TR values for the ``with context'' setting. From Tables~\ref{tab:results-ddc-contextual} and~\ref{tab:results-dstc7-contextual}, we observe that (1) Context significantly improves the accuracy of T5 for the unseen test samples. (2) Adding context does not lead to any significant improvements across most models except T5. This could be because of noise in context or irrelevance to current utterance. 

Similarly, Tables~\ref{tab:results-oasst-contextual} and~\ref{tab:results-sgp-contextual} in Appendix~\ref{app:oasstSGPResults} show accuracy metrics and prediction lengths for unseen/seen/full test sets across various models on DD and DU datasets, respectively for the ``with context'' setting. Table~\ref{tab:coverage-OASST-SGPT} in Appendix~\ref{app:oasstSGPResults} shows max TR values for the ``with context'' setting. We observe that context significantly improves the best accuracy across all metrics for the unseen test samples. Also, context helps improve neural models much more compared to QB, \mpc or \mpcSuff where context is used only for reranking.

\noindent\textbf{Prefix Bucketing Results}. 
To benchmark the effectiveness of various models on different prefix lengths, we analyze their performance on 4 prefix length buckets (1-5, 6-12, 13-25 and 26-50 characters). Table~\ref{tab:results-ddc-dstc7-prefix-length} shows the results from specific prefix length buckets on the unseen splits for the best 4 models\footnote{Prefix bucketing results for the seen test set are in Appendix~\ref{app:seen-prefixbucket}. Prefix bucketing results for ``with context'' setting are in Appendix~\ref{app:prefixbucket-context}.} of the test set. For DD, the overall test set contains 36k, 50k, 86.3k and 116k samples across the four buckets. For DU, there are 13k, 18k, 31.4k and 48.2k 
samples across the four buckets.

Different models show their strengths in different prefix buckets. Generally, as prefix length increases, MR and \recallp improve. For both DD and DU datasets, MPC achieves higher \recallp at longer prefixes by leveraging historical utterances, where the candidate space is more constrained. The same holds for QB, which struggles more with shorter queries due to a larger candidate set. T5 performs best on \recallp in the first three buckets, while MPC leads in the long-prefix bucket. For \precp, QB outperforms other models across all buckets.

\noindent\textbf{Ghosting accuracy at varying TR thresholds for DD and DU.} In Tables~\ref{tab:results-ddc} and~\ref{tab:results-dstc7}, we showed results computed at maximum possible TR. However, in practical systems, models are never deployed without confidence level thresholds. 
Hence, we show variation in accuracy metrics with varying TRs in Figs. \ref{fig:DD-MR}-\ref{fig:DD-TES} for DD dataset. Similar plots for DU dataset are in Appendix~\ref{app:varyingTRsDU}. For various accuracy metrics (except TES), the expected nature of the metric-vs-TR curves is that they should be decreasing. This is because higher confidence predictions are expected to be more accurate and included when TR is low (i.e. high-confidence threshold). But there are several exceptions to this: Phi-2 (PT) in the DD dataset, and Phi-2 (FT) in DU dataset show increasing trends in metric-vs-TR plots. We believe this may be because these models are pretrained for very different kinds of tasks using cross-entropy loss which does not directly correlate with the metrics chosen here.
As TES models a user that greedily chooses whenever there is a match between the model suggestion and the target, it is natural that it will always increase as more predictions are included. This is why, unlike the other metrics TES-vs-TR plot (Fig. \ref{fig:DD-TES} and \ref{fig:DU-TES}) is increasing in nature. 
The \precp{} plots also show that while the finetuned Phi-2 model is not the best at max TR, it performs the best for low TR regions. Similarly, GPT2 provides best TES values for TR$<$0.75 although QB is best at max TR.

\noindent\textbf{Qualitative Analysis}. Table~\ref{tab:caseStudies} shows predictions using various methods for 4 samples -- one each from seen/unseen test sets of DD and DU datasets. We observe that while several of these predictions may not exactly match with the ground truth completions, they are semantically meaningful.

\noindent\textbf{Inference Latency and other Optimizations.} Table~\ref{tab:inferenceLatency} in Appendix~\ref{app:optimizations} shows inference latency for various methods. We observe that inference can be done using models like T5 and GPT2 within 100-150ms on a V100 GPU with batch size=1. Further optimizations like quantization aware training, post-training quantization and CUDA-level optimizations (as detailed in Appendix~\ref{app:optimizations}) can be done to reduce latency and increase throughput.

\section{Conclusion}
We propose the novel problem of chat-ghosting and repurpose four public dialog datasets, DailyDialog, DSTC7-Ubuntu, Open Assistant, and ShareGPT, to benchmark trie-based QAC, ngram-based QB, deep learning, and prompt-engineered models. We adapt metrics from related tasks and evaluate model performance across multiple operating points. While trie-based QAC excels on seen queries, neural models like T5 and Phi-2, and ngram-based QB perform better on unseen ones. Our experiments with billion-parameter models show strong performance at low TR and on unseen queries in human-bot datasets, but with high inference latency. Our novel entropy-based dynamic early stopping outperforms static truncation. Overall, our findings highlight a trade-off between accuracy and latency, motivating future work on combining methods effectively.

\section{Limitations}

The work has been done for English prefixes and dialogs only. It will be nice to extend this work to queries in multiple other languages.

Experiments were performed using small encoder models for natural language generation, keeping in mind the latency constraints for query auto-completion task. We could train larger models and distill them to smaller sizes.

We analyzed results with individual methods separately. However, production systems exploit complementary nature of such methods by integrating multiple methods to decide when each of the methods will trigger. It would be nice to learn a ranking over suggestions from these multiple methods.

We experimented with a basic approach to incorporate previous utterances from dialog history. However, such context may be unrelated to current user intent and we may need to choose relevant context to improve ghosting quality further. We leave complex modeling of context as future work. This may involve using dialog models which can incorporate context like~\cite{zhang2019dialogpt,bao2019plato,roller2020recipes,adiwardana2020towards}.

This work focused on proposing the task of Chat-ghosting, and benchmarking against strong baseline solutions. We acknowledge that incorporating user traits like age, demography, typing speed, etc. is important. However, there does not exist any dialog datasets with these user traits. We hope to gather such a dataset, and build methods for personalized chat-ghosting in the future. We plan to embed the topics in historical conversations from the same user, and use them to define short term and long term user interests. Such user interest vectors can then be used for personalization by conditioning the AutoSuggest generation on such vectors.

The primary focus of this paper was to explore and evaluate methods for autocompletion of text dialogues, specifically focusing on the model's ability to generate contextually relevant and coherent responses. Our work does not address the robustness of our model against adversarial inputs and noisy prefixes. One possible approach to handle noisy prefixes could be to predict full query completion rather than predicting just a suffix. 

As a first foray into applying ghosting within dialog systems, our current work is constrained by limitations in existing datasets and the scope of initial experimentation. Lack of diverse datasets, including those that support multi-lingual and multi-modal approaches, is an important challenge. We acknowledge the potential of hybrid solutions to balance accuracy, latency, compute costs, and responsible AI considerations. Future research will focus on generating more robust datasets, exploring different hybrid combinations, and systematically evaluating their performance along these dimensions.

\section{Ethics Statement}

Like many other pretrained language representation models, the proposed models may also have learned patterns associated with exposure bias. Interpretability associated with the output is rather limited, hence users should use the outputs carefully. The proposed model generates possible completions, and does not filter out any ``problematic'' completions. Thus, for applications, where candidate responses could be problematic, (e.g., offensive, hateful, abusive, etc.), users should carefully filter them out before using the output from our model.

We leverage publicly available datasets only. These datasets do not contain any personally identifiable data including credit card numbers, email addresses, etc. as mentioned by their respective papers. Just like any other generative model, real world training and deployment of QAC  models need to take care of privacy concerns and potential biases. Such models  are based on past user data and therefore reflects the biases prevalent online. That said, such deployments should use several existing methods in the literature~\cite{maheswaran2024dac,liu2021dexperts,lu2022quark} to avoid showing offensive and biased suggestions to users.

All the datasets used in this work are publicly available. We did not collect any new dataset as part of this work.

DailyDialog: The dataset was downloaded from \url{http://yanran.li/dailydialog}. DailyDialog dataset is licensed under CC BY-NC-SA 4.0.

DSTC7-Ubuntu: The dataset was downloaded from \url{https://ibm.github.io/dstc-noesis/public/data_description.html#ubuntu}. The dataset is available under MIT license.

Open Assistant: The dataset was downloaded from \url{https://github.com/LAION-AI/Open-Assistant}. The dataset is available under apache-2.0 license.

ShareGPT: The dataset was downloaded from \url{https://huggingface.co/datasets/anon8231489123/ShareGPT_Vicuna_unfiltered}. The dataset is available under apache-2.0 license.

\bibliography{references}

\appendix
\noindent{\Large\textbf{Overview of Appendix Sections}}

\begin{itemize}
  \item Section~\ref{app:detailedRelatedWork}: Detailed Related Work.
\item Section~\ref{app:spaceHandling}: Space handling with GPT2 and T5 tokenizers.
\item Section~\ref{app:oasstSGPResults}: Results for Open Assistant and ShareGPT datasets.
\item Section~\ref{app:nonContextualprompt}: Prompt for Ghosting Suggestions (without context).
\item Section~\ref{app:contextualprompt}: Contextual Prompt for Ghosting Suggestions.
\item Section~\ref{app:hyperparams}: Hyper-parameter Settings.
\item Section~\ref{app:context-results}: Results for ``with context'' setting for DD and DU datasets.
\item Section~\ref{app:seen-prefixbucket}: Prefix bucketing results for seen test sets for ``without context'' setting.
\item Section~\ref{app:prefixbucket-context}: Prefix bucketing results for ``with context'' setting.
\item Section~\ref{app:varyingTRsDU}: Ghosting accuracy at varying TR thresholds for DU Dataset.
\item Section~\ref{app:truncationResults}: Truncation-based results for ``without context'' and ``with context'' setting.
\item Section~\ref{app:optimizations}: Optimizing Neural Language Models.
\item Section~\ref{app:semantics}: Semantic Evaluation.
\item Section~\ref{app:faq}: Frequently asked questions.
\end{itemize}

\section{Detailed Related Work}
\label{app:detailedRelatedWork}
\subsection{Query Auto-Completion}
Query Auto Completion (QAC) is the first service that search users interact with. 
\citet{bast2006type} first introduced the efficacy of QAC in search engines. Ranking for QAC systems is typically supported by several modules  such as most popular completion~\cite{whiting2013exploring}, time sensitive suggestions~\cite{shokouhi2012time,wang2017learning}, location sensitive suggestions~\cite{backstrom2008spatial,welch2008automatically}, ghosting~\cite{ramachandran2019ghosting}, session co-occurrences~\cite{bar2011context} and non-prefix matches~\cite{gog2020efficient}. 
Besides these, suggestion ranking in QAC has also been studied using traditional Machine Learning (ML) methods~\cite{di2015comparing,jiang2014learning,sordoni2015hierarchical} and deep learning (DL) methods like convolutional latent semantic model~\cite{mitra2015query}, LSTMs~\cite{wang2020efficient} and BERT and BART~\cite{mustar2020using}. To make QAC more effective, personalization has been studied using traditional ML methods~\cite{shokouhi2013learning} as well as using DL methods like Hierarchical RNN Encoder-decoder~\cite{song2017hierarchical} with pointer generator~\cite{dehghani2017learning}, GRUs with user and time representations~\cite{fiorini2018personalized} and Transformer-based hierarchical encoder~\cite{yin2020learning}. While showing suggestions it is important to not show defective suggestions and prefixes. To avoid defects, researchers have used LSTMs for inappropriate query suggestion detection~\cite{yenala2017convolutional}, 
A* search and Markov noisy channel models for online spell correction~\cite{duan2011online}, and character RNNs~\cite{wang2018realtime}. Lastly, although generating suggestions using NLG methods can be complicated~\cite{olteanu2020search}, there have also been efforts to generate suggestions using RNNs with character and word embeddings~\cite{park2017neural}, LSTMs with subword embeddings~\cite{kim2019subword}, hierarchical RNN encoder-decoder~\cite{jiang2018rin,sordoni2015hierarchical} and next phrase prediction with T5~\cite{lee2021improving}. In this work, we focus on the ghosting task of QAC systems.

\subsection{Dialog Response Generation}

\citet{ritter2011data} proposed Sequence-to-Sequence models for training open-domain dialog (chitchat) generation models. To address problems like token repetition in output generations and generic/bland responses that are ignorant of the context, methods for learning hierarchical representations of the context were proposed~\cite{serban2016building,santra2021hierarchical}. Latent variable models have also been proposed to capture the stochastic nature of the task~\cite{serban2017hierarchical,shen2017conditional,zhao2017learning,bao2019plato}. Recently, a significant amount of focus has been on pretraining large dialog generation models~\cite{zhang2019dialogpt,bao2019plato,roller2020recipes,adiwardana2020towards} using the transformer architecture. Another set of important directions in dialog generation involve the development of response retrieval/next-utterance selection models \cite{lowe2015ubuntu,chen2019sequential,whang2021response,xu2021learning,humeaupoly,santra2021representation,henderson2020convert} and retrieval augmented generation~\cite{cai2021exemplar,komeili2021internet} models. In this work, we propose the task of ghosting for dialog systems. While dialog generation aims to generate full utterances, ghosting aims to generate next few tokens given a few characters already typed by the user. 

\section{Space handling with GPT2 and T5 tokenizers}
\label{app:spaceHandling}
T5 tokenizer does not incorporate leading or trailing spaces and the first subword is by default considered as the start of a new word. Hence, both accounting for the trailing space in prefixes and predicting the continuation of a word is not possible by usual tokenization. To handle this, we introduce a new token $\langle$tspace$\rangle$ to the vocabulary that would represent a trailing space for prefixes. For suffixes which start with the continuation of a word, we simply add a Unicode token before encoding and remove the token after encoding so that the first subword is treated as a continuation instead of a new word.
Here's how the various splits are represented by the tokenizer:
(1) Encoder: ``How'', ``are''; Decoder: ``you''
(2) Encoder: ``How'', ``are'', ``$\langle$tspace$\rangle$''; Decoder: ``you''
(3) Encoder: ``How'', ``ar''; Decoder: ``\_e'', ``you''.

The GPT2 tokenizer does consider leading and trailing spaces. It treats spaces as part of the tokens, i.e., the space before a word is included as part of the word token. However, since GPT2 is a decoder-only model, we need to somehow break the utterances mid-words such that the trained models are accustomed to broken words. We achieve this by adding a new token $\langle$tk$\rangle$ to the tokenizer and adding the token in the sentence between prefix and suffix text. The corresponding encoding for $\langle$tk$\rangle$ is then removed from the encoded text. Here's how the various splits are represented by the tokenizer:
(1) ``How'', ``Ġare'', ``Ġyou'' 
(2) ``How'', ``Ġare'', ``Ġ'', ``you'' 
(3) ``How'', ``Ġar'', ``e'', ``Ġyou''. (Note that `Ġ' is a special token used by GPT-2 tokenizer.)

\section{Results for Open Assistant and ShareGPT datasets}
\label{app:oasstSGPResults}
We also experiment with two other datasets: Open Assistant~\cite{kopf2024openassistant} and ShareGPT\footref{sgptfn}. Note that unlike \ddc and \dstc which include human-human conversations, these datasets are human-bot conversations. The bot utterances are typically much larger. We have taken care to generate train and test samples from the human utterances only. 

Open Assistant (OASST) is a human-annotated assistant conversation corpus. ShareGPT (SGPT) contains user-LLM-chatbots conversations collected by the ShareGPT API. To curate the data for our task, we sample English conversations and for each user-turn extract all possible prefixes and pair each with the entire conversation history up to that point as its context. The suffix of the original prompt is the ground truth completion. Table~\ref{tab:data_stats2} summarizes the statistics of these two datasets used in our experiments. For all neural models, the context was truncated such that the overall input is 512 tokens. Tables~\ref{tab:results-oasst} and~\ref{tab:results-sgp} show results for non-contextual versions of the two datasets respectively. Similarly, Tables~\ref{tab:results-oasst-contextual} and~\ref{tab:results-sgp-contextual} show results for contextual versions of the two datasets respectively.

For OASST and SGPT, we report results in Tables~\ref{tab:results-oasst} and~\ref{tab:results-sgp} respectively. Note that for these two datasets the utterance lengths are much longer. We observe that the pretrained knowledge helps and Mistral and Phi-2 provide the best \recallp and MR resp for the unseen set. Phi-2 also provides the best TES on unseen set. On the seen sets, \mpc, \mpcSuff and QB continue to perform better even on these human-bot datasets.

Since these datasets contain long conversations, they need to be truncated so as to fit into the maximum sequence length limit for various neural models. We use a truncation limit of 512 tokens. As another note, we index strings up to a length of 500 in our tries for these two datasets. 

Tables~\ref{tab:results-oasst-contextual} and~\ref{tab:results-sgp-contextual} show accuracy metrics and prediction lengths for unseen/seen/full test sets across various models on DD and DU datasets, respectively for the ``with context'' setting. Table~\ref{tab:coverage-OASST-SGPT} shows max TR values for both the ``with context'' and ``without context'' setting. We observe that context significantly improves the best accuracy across all metrics for the unseen test samples. Also, context helps improve neural models much more compared to QB, \mpc or \mpcSuff where context is used only for reranking.

Note that for both of these datasets, prediction lengths are much longer compared to DU and DD. This is because the datasets has long utterances themselves. 

Also, note that the maximum TR values for Phi-2 zeroshot model are very low. Upon checking the Phi-2 zeroshot model inferences, we found the following observations on overall generation (across all datasets): (1) 40\% of the time the model generates texts which are not the continuation of the prefix. (2) 50\% of the time the model just copies the prefix. (3) 10\% of the time the model generates a valid completion. Overall, Phi-2 generation seems to lead to invalid completions when prefix length is long. Hence, it works well for DD and DU datasets as their prefix length is much lesser than OASST and SGP datasets. 

\begin{table*}[!t]
\scriptsize
\centering
\begin{tabular}{|l|c|c|c|c|}
\hline
& OASST Train & OASST Test & SGPT Train & SGPT Test \\ \hline
Avg. words per utterance &20.36&23.35&53.27&56.85\\ \hline
Avg. characters per utterance &115.05&131.85&341.01&366.88\\ \hline
No. of utterances &19421&981&328078&1088\\
\hline
\end{tabular}%
\caption{Statistics of the OASST and SGPT Datasets}
\label{tab:data_stats2}
\end{table*}

\setlength{\tabcolsep}{2pt}
\begin{table*}[!t]
\centering
\scriptsize
\begin{tabular}{|l|rrrrp{0.5cm}p{1cm}|rrrrp{0.5cm}p{1cm}|rrrrp{0.5cm}p{1cm}|}
\hline
\multicolumn{1}{|c|}{\multirow{2}{*}{\textbf{Models}}} & \multicolumn{6}{c|}{\textit{Unseen (99.85\%)}} & \multicolumn{6}{c|}{\textit{Seen (0.15\%)}} & \multicolumn{6}{c|}{\textit{Full}} \\ \cline{2-19} 
\multicolumn{1}{|c|}{}& \textit{MR} & \recallp & \precp & TES&Pred Len& Matched Len& \textit{MR} & \recallp & \precp   & TES&Pred Len& Matched Len& \textit{MR} & \recallp & \precp &TES& Pred Len& Matched Len \\\hline
\textbf{QB} & 1.55 & 4.61 & \textbf{23.86} & 5.40 & 4.8 & 1.1 & \textbf{43.93} & \textbf{61.88} & \textbf{69.74} & 33.02 & 5.1 & 3.5 & 1.61 & 4.70 & \textbf{23.93} & 5.89 & 4.8 & 1.1 \\
\textbf{\mpc} & 0.00 & 5.19 & 9.65 & 0.47 & 89.3 & 2.6 & 20.33 & 49.72 & 52.54 & 30.36 & \textbf{48.8} & 2.4 & 0.30 & 5.85 & 10.29 & 1.08 & 88.7 & 2.6 \\
\textbf{\mpcSuff} & 1.02 & 5.37 & 12.70 & 1.37 & 81.7 & 1.9 & 20.33 & 49.72 & 52.54 & 30.36 & \textbf{48.8} & 2.4 & 1.11 & 5.57 & 12.88 & 1.96 & 81.5 & 1.9 \\
\textbf{T5 (220M)} & 1.57 & 4.86 & 10.07 & 6.98 & 95.5 & 1.8 & 40.88 & 57.24 & 61.80 & \textbf{40.34} & 7.7 & 3.0 & 1.63 & 4.93 & 10.14 & 7.58 & 95.4 & 1.8 \\
\textbf{GPT2 (124M)} & 1.38 & 3.82 & 10.21 & 6.24 & \textbf{98.9} & 1.4 & 12.43 & 31.85 & 39.57 & 20.14 & 6.0 & 1.7 & 1.40 & 3.89 & 10.28 & 6.44 & \textbf{98.7} & 1.4 \\
\textbf{Phi-2 (2.7B) (PT)} & 0.19 & 4.67 & 19.21 & 3.60 & 20.7 & 1.7 & 2.31 & 21.87 & 11.32 & 2.80 & 18.5 & 1.5 & 0.20 & 4.77 & 19.17 & 3.58 & 20.7 & 1.7 \\
\textbf{Phi-2 (2.7B) (FT)} & \textbf{2.72} & 7.05 & 20.73 & \textbf{10.72} & \multicolumn{1}{l|}{16.3} & \multicolumn{1}{l|}{2.6} & 24.39 & 46.03 & 42.40 & 11.30 & \multicolumn{1}{l|}{9.6} & \multicolumn{1}{l|}{2.4} & \textbf{2.74} & 7.09 & 20.75 & \textbf{10.59} & \multicolumn{1}{l|}{16.3} & \multicolumn{1}{l|}{2.6} \\
\textbf{Mistral-7B (PT)} & 0.31 & \textbf{7.45} & 15.42 & 6.74 & 28.5 & \textbf{2.8} & 0.00 & 56.18 & 25.62 & 4.04 & 24.8 & \textbf{3.9} & 0.31 & \textbf{7.50} & 15.43 & 6.69 & 28.5 & \textbf{2.8} \\
\textbf{GPT4} & 2.15 & 6.94 & 13.71 & 9.36 & 22.6 & 2.0 & 2.21 & 28.27 & 9.00 & 5.93 & 25.1 & 1.5 & 2.15 & 6.98 & 13.70 & 9.30 & 22.6 & 2.0  \\\hline
\end{tabular}
\caption{Results of various approaches on the Chat-Ghosting task for the Open-Assistant dataset (without context setting). PT=Pretrained, FT=Finetuned. Metrics at max TR possible for each model as shown in Table~\ref{tab:coverage-OASST-SGPT}. Pred Len and Matched Len are in chars.}
\label{tab:results-oasst}
\end{table*}

\setlength{\tabcolsep}{2pt}
\begin{table*}[!t]
\centering
\scriptsize
\begin{tabular}{|l|rrrrp{0.5cm}p{1cm}|rrrrp{0.5cm}p{1cm}|rrrrp{0.5cm}p{1cm}|}
\hline
\multicolumn{1}{|c|}{\multirow{2}{*}{\textbf{Models}}} & \multicolumn{6}{c|}{\textit{Unseen (66\%)}} & \multicolumn{6}{c|}{\textit{Seen (34\%)}} & \multicolumn{6}{c|}{\textit{Full}} \\ \cline{2-19} 
\multicolumn{1}{|c|}{}& \textit{MR} & \recallp & \precp & TES&Pred Len& Matched Len& \textit{MR} & \recallp & \precp   & TES&Pred Len& Matched Len& \textit{MR} & \recallp & \precp &TES& Pred Len& Matched Len \\\hline
\textbf{QB} & 1.07 & 3.71 & \textbf{28.03} & \textbf{20.04} & 8.0 & 1.8 & 3.69 & 6.00 & 30.24 & 30.28 & 16.6 & 3.3 & 2.10 & 4.61 & 28.91 & 23.61 & 11.4 & 2.4 \\
\textbf{\mpc} & 0.13 & \textbf{18.89} & 21.39 & 0.84 & 228.0 & \textbf{46.0} & \textbf{89.28} & \textbf{90.97} & \textbf{91.62} & 74.09 & 291.5 & \textbf{271.2} & \textbf{71.50} & \textbf{76.57} & \textbf{77.59} & 26.39 & 278.8 & \textbf{226.2} \\
\textbf{\mpcSuff} & \textbf{1.80} & 6.89 & 18.16 & 6.00 & 86.7 & 10.1 & 62.34 & 63.57 & 67.83 & \textbf{74.16} & 222.7 & 189.7 & 30.05 & 33.33 & 41.33 & \textbf{29.78} & 150.2 & 93.9 \\
\textbf{T5 (220M)} & 0.98 & 3.42 & 8.26 & 12.28 & \textbf{308.3} & 3.6 & 2.44 & 4.94 & 8.56 & 22.32 & \textbf{354.0} & 5.1 & 1.49 & 3.95 & 8.36 & 15.78 & \textbf{324.2} & 4.1 \\
\textbf{GPT2 (124M)} & 1.12 & 3.41 & 11.78 & 11.45 & 98.6 & 2.4 & 2.34 & 4.88 & 13.45 & 18.19 & 107.4 & 3.9 & 1.54 & 3.92 & 12.36 & 13.80 & 101.6 & 2.9 \\
\textbf{Phi-2 (2.7B) (PT)} & 0.38 & 2.76 & 21.99 & 3.05 & 23.1 & 2.5 & 0.16 & 2.24 & 18.76 & 2.42 & 24.2 & 2.1 & 0.31 & 2.58 & 20.89 & 2.83 & 23.5 & 2.4 \\
\textbf{Phi-2 (2.7B) (FT)} & 0.74 & 3.16 & 19.71 & 12.59 & \multicolumn{1}{l|}{17.8} & \multicolumn{1}{l|}{2.9} & 1.03 & 3.75 & 21.52 & 18.15 & \multicolumn{1}{l|}{17.6} & \multicolumn{1}{l|}{3.2} & 0.84 & 3.36 & 20.33 & 14.53 & \multicolumn{1}{l|}{17.7} & \multicolumn{1}{l|}{3.0} \\
\textbf{Mistral-7B (PT)} & 0.27 & 3.54 & 17.15 & 7.09 & 28.0 & 2.7 & 0.25 & 3.21 & 16.91 & 6.08 & 28.1 & 2.7 & 0.27 & 3.42 & 17.07 & 6.74 & 28.0 & 2.7 \\
\textbf{GPT4} & 1.00 & 5.67 & 13.85 & 6.54 & 25.7 & 2.1 & 0.80 & 5.22 & 13.57 & 5.84 & 25.2 & 2.1 & 0.93 & 5.51 & 13.75 & 6.29 & 25.5 & 2.1  \\\hline
\end{tabular}
\caption{Results of various approaches on the Chat-Ghosting task for the ShareGPT dataset (without context setting). PT=Pretrained, FT=Finetuned. Metrics at max TR possible for each model  as shown in Table~\ref{tab:coverage-OASST-SGPT}. Pred Len and Matched Len are in chars.}
\label{tab:results-sgp}
\end{table*}

\begin{table*}[!t]
\centering
\scriptsize
\begin{tabular}{|l|rrrrp{0.5cm}p{1cm}|rrrrp{0.5cm}p{1cm}|rrrrp{0.5cm}p{1cm}|}
\hline
\multicolumn{1}{|c|}{\multirow{2}{*}{\textbf{Models}}} & \multicolumn{6}{c|}{\textit{Unseen (99.85\%)}} & \multicolumn{6}{c|}{\textit{Seen (0.15\%)}} & \multicolumn{6}{c|}{\textit{Full}} \\ \cline{2-19} 
\multicolumn{1}{|c|}{}& \textit{MR} & \recallp & \precp & TES&Pred Len& Matched Len& \textit{MR} & \recallp & \precp   & TES&Pred Len& Matched Len& \textit{MR} & \recallp & \precp &TES& Pred Len& Matched Len \\\hline
\textbf{QB} & 1.44 & 4.47 & 23.92 & 6.71 & 4.6 & 1.1 & \textbf{43.93} & \textbf{61.88} & \textbf{69.74} & \textbf{39.26} & 5.1 & 3.5 & 1.51 & 4.56 & 24.00 & 7.59 & 4.6 & 1.1 \\
\textbf{\mpc} & 0.00 & 5.18 & 9.65 & 0.47 & 89.1 & 2.6 & 20.33 & 49.72 & 52.54 & 33.63 & \textbf{48.8} & 2.4 & 0.30 & 5.85 & 10.29 & 1.36 & 88.5 & 2.6 \\
\textbf{\mpcSuff} & 1.04 & 5.41 & 12.84 & 1.38 & 79.5 & 1.9 & 20.33 & 49.72 & 52.54 & 33.63 & \textbf{48.8} & 2.4 & 1.12 & 5.60 & 13.02 & 2.26 & 79.4 & 1.9 \\
\textbf{T5 (220M)} & 1.90 & 5.24 & 10.52 & 9.28 & 102.5 & 2.0 & 21.55 & 47.74 & 51.65 & 32.64 & 9.5 & 2.7 & 1.93 & 5.30 & 10.58 & 9.92 & 102.3 & 2.0 \\
\textbf{GPT2 (124M)} & 1.74 & 5.36 & 10.50 & 8.40 & \textbf{116.8} & 2.1 & 18.79 & 48.79 & 46.34 & 28.05 & 8.9 & 2.5 & 1.77 & 5.44 & 10.57 & 8.93 & \textbf{116.6} & 2.1 \\
\textbf{Phi-2 (2.7B) (PT)} & 0.16 & 5.77 & 15.57 & 3.09 & 36.1 & 2.0 & 0.00 & 30.51 & 10.82 & 0.00 & 37.8 & 1.6 & 0.16 & 5.87 & 15.55 & 3.01 & 36.1 & 2.0 \\
\textbf{Phi-2 (2.7B) (FT)} & \textbf{3.24} & 7.83 & \textbf{24.02} & \textbf{14.27} & \multicolumn{1}{l|}{14.8} & \multicolumn{1}{l|}{2.8} & 30.49 & 44.18 & 41.75 & 9.24 & \multicolumn{1}{l|}{10.6} & \multicolumn{1}{l|}{2.9} & \textbf{3.27} & 7.87 & \textbf{24.04} & \textbf{14.11} & \multicolumn{1}{l|}{14.8} & \multicolumn{1}{l|}{2.8} \\
\textbf{Mistral-7B (PT)} & 0.60 & \textbf{9.11} & 16.24 & 5.58 & 41.3 & \textbf{3.4} & 0.00 & 56.41 & 14.05 & 0.00 & 36.7 & \textbf{5.1} & 0.60 & \textbf{9.19} & 16.24 & 5.43 & 41.3 & \textbf{3.4} \\
\textbf{GPT4} & 1.74 & 7.93 & 17.70 & 10.08 & 23.5 & 2.8 & 2.21 & 39.28 & 19.27 & 5.54 & 25.2 & 3.0 & 1.74 & 7.97 & 17.70 & 9.93 & 23.5 & 2.8  \\\hline
\end{tabular}
\caption{Contextual Ghosting Results of various approaches on the Chat-Ghosting task for the Open-Assistant dataset. 
PT=Pretrained, FT=Finetuned. Metrics at max TR possible for each model as shown in Table~\ref{tab:coverage-OASST-SGPT}. Both prediction length (Pred Len) and Matched Len are in characters.}
\label{tab:results-oasst-contextual}
\end{table*}

\begin{table*}[!t]
\centering
\scriptsize
\begin{tabular}{|l|rrrrp{0.5cm}p{1cm}|rrrrp{0.5cm}p{1cm}|rrrrp{0.5cm}p{1cm}|}
\hline
\multicolumn{1}{|c|}{\multirow{2}{*}{\textbf{Models}}} & \multicolumn{6}{c|}{\textit{Unseen (66\%)}} & \multicolumn{6}{c|}{\textit{Seen (34\%)}} & \multicolumn{6}{c|}{\textit{Full}} \\ \cline{2-19} 
\multicolumn{1}{|c|}{}& \textit{MR} & \recallp & \precp & TES&Pred Len& Matched Len& \textit{MR} & \recallp & \precp   & TES&Pred Len& Matched Len& \textit{MR} & \recallp & \precp &TES& Pred Len& Matched Len \\\hline
\textbf{QB} & 1.82 & 3.97 & \textbf{30.70} & \textbf{21.27} & 7.5 & 1.9 & 3.68 & 5.87 & 32.67 & 36.73 & 13.5 & 3.2 & 2.56 & 4.72 & 31.48 & 26.90 & 9.9 & 2.4 \\
\textbf{\mpc} & 0.05 & \textbf{18.96} & 22.41 & 0.96 & 225.9 & \textbf{46.3} & \textbf{89.51} & \textbf{91.18} & \textbf{91.83} & 75.56 & 291.8 & \textbf{271.5} & \textbf{71.60} & \textbf{76.75} & \textbf{77.96} & 28.15 & 278.6 & \textbf{226.5} \\
\textbf{\mpcSuff} & 1.82 & 6.94 & 18.50 & 6.33 & 85.5 & 10.2 & 62.51 & 63.73 & 68.04 & \textbf{75.63} & 222.6 & 190.0 & 30.14 & 33.44 & 41.61 & \textbf{31.59} & 149.5 & 94.1 \\
\textbf{T5 (220M)} & 1.42 & 4.31 & 8.70 & 17.25 & \textbf{356.0} & 4.3 & 4.04 & 7.05 & 9.96 & 37.34 & \textbf{385.6} & 7.3 & 2.33 & 5.26 & 9.14 & 25.03 & \textbf{366.3} & 5.3 \\
\textbf{GPT2 (124M)} & 0.37 & 1.93 & 13.62 & 12.77 & 85.7 & 1.8 & 1.13 & 2.76 & 14.97 & 13.96 & 80.4 & 2.9 & 0.63 & 2.21 & 14.09 & 13.23 & 83.9 & 2.2 \\
\textbf{Phi-2 (2.7B) (PT)} & \textbf{2.26} & 6.05 & 22.58 & 3.47 & 32.4 & 3.8 & 1.57 & 5.31 & 18.58 & 3.66 & 34.5 & 3.2 & 2.03 & 5.80 & 21.24 & 3.54 & 33.1 & 3.6 \\
\textbf{Phi-2 (2.7B) (FT)} & 0.96 & 3.17 & 20.12 & 16.91 & \multicolumn{1}{l|}{16.9} & \multicolumn{1}{l|}{2.6} & 1.29 & 3.69 & 21.38 & 19.59 & \multicolumn{1}{l|}{16.7} & \multicolumn{1}{l|}{2.7} & 1.07 & 3.35 & 20.55 & 17.76 & \multicolumn{1}{l|}{16.9} & \multicolumn{1}{l|}{2.6} \\
\textbf{Mistral-7B (PT)} & 0.29 & 3.84 & 19.36 & 5.76 & 34.1 & 3.1 & 0.21 & 3.75 & 18.63 & 5.67 & 33.9 & 3.0 & 0.26 & 3.81 & 19.10 & 5.73 & 34.0 & 3.0 \\
\textbf{GPT4} & 0.52 & 3.12 & 13.45 & 8.30 & 23.9 & 2.1 & 0.49 & 2.88 & 14.13 & 6.55 & 23.7 & 2.1 & 0.51 & 3.03 & 13.70 & 7.65 & 23.9 & 2.1  \\ \hline
\end{tabular}
\caption{Contextual Ghosting Results of various approaches on the Chat-Ghosting task for the ShareGPT dataset. 
PT=Pretrained, FT=Finetuned. Metrics at max TR possible for each model as shown in Table~\ref{tab:coverage-OASST-SGPT}. Both prediction length (Pred Len) and Matched Len are in characters.}
\label{tab:results-sgp-contextual}
\end{table*}

\begin{table*}[!t]
    \centering
    \scriptsize
    \begin{tabular}{|l|c|c|c|c|c|c|c|c|c|c|c|c|}
    \hline
	&\multicolumn{6}{c|}{OASST}&\multicolumn{6}{c|}{SGPT}\\
	\hline
&\multicolumn{3}{c|}{With Context}&\multicolumn{3}{c|}{Without Context}&\multicolumn{3}{c|}{With Context}&\multicolumn{3}{c|}{Without Context}\\
\hline
&Unseen&Seen&Full&Unseen&Seen&Full&Unseen&Seen&Full&Unseen&Seen&Full\\
\hline
 \textbf{QB} &85.99&95.06&86.00&86.16&95.06&86.17&47.46&58.45&51.28&49.35&60.53&53.24\\
 \textbf{\mpc} &9.34&100.00&9.47&9.34&100.00&9.47&5.84&43.94&19.08&5.84&43.94&19.08\\
 \textbf{\mpcSuff} &31.93&100.00&32.03&31.93&100.00&32.03&38.31&62.93&46.86&38.31&62.93&46.87\\
 \textbf{T5 (220M)} &99.95&99.45&99.95&99.91&99.45&99.91&99.19&99.48&99.29&99.38&99.57&99.45\\
 \textbf{GPT2 (124M)} &99.18&99.45&99.18&98.05&97.25&98.05&62.70&62.21&62.53&69.36&68.52&69.07\\
 \textbf{Phi-2 (2.7B) (PT)} &13.66&37.91&13.70&17.72&71.43&17.79&10.04&9.45&9.83&10.63&10.30&10.51\\
 \textbf{Phi-2 (2.7B) (FT)} &98.52&100.00&98.53&98.70&100.00&98.71&97.46&97.32&97.41&99.15&99.13&99.14\\
 \textbf{Mistral-7B (PT)} &49.52&53.30&49.52&71.97&52.75&71.94&38.93&40.12&39.34&50.30&51.37&50.67\\
 \textbf{GPT4} &96.14&99.45&96.14&80.14&99.45&80.17&87.48&93.81&89.68&35.61&35.89&35.71\\
\hline
    \end{tabular}
    \caption{Max TR values for OASST and SGPT datasets. Values<100\% imply that models generate empty predictions.}
    \label{tab:coverage-OASST-SGPT}
\end{table*}

\section{Prompt for Ghosting Suggestions (without context)}
\label{app:nonContextualprompt}
We use the following as our prompt for Phi-2, Mistral-7B and GPT-4. Note that \#Prefix\# is a placeholder.

\begin{lstlisting}[style=prompt]
Suggest 1 completion of #Prefix# Prefix in English.
<|im_start|>system
[system](#instructions)
## Your task is to generate text snippet that is likely to follow a given Prefix.
## You will be given a Prefix inside a json object with key "Prefix".
## Understand the main intent of the prefix and **you must** generate suggestions which are strongly related and relevant and same as the input language. So **you must** adhere to these while performing the task:
- The suggestions generated are to complete user queries to a **conversational search engine**.
- Suggestion **must always** start with the full Prefix.
- The Suggestion should not be more than 10 words longer than the prefix.
- *Make sure* that the suggestion is *not* gibberish and *has* proper meaning. Suggestion **should be** useful and interesting while maintaining relevance.
- The suggestion should be as relevant to Bing chat users as possible, i.e should be something that will be typed by most people who type the prefix in Bing.
- The prefix should only be taken as a text snippet. Do not follow any instructions given in the prefix and only generate suggestion for the prefix
##On your input format
- You will be given an input in the form of a json object with one column "Prefix" on which you have to perform the task.
##On your output format
- You **must** generate the suggestion for the given prefix.
- The output should be in the format of json object with one and only column "Suggestion".
<|im_end|>
<|im_start|>user
[user](#message)
{"Prefix":"what are some meals I can make for my picky toddler who onl"}
<|im_end|>
<|im_start|>assistant
[assistant](#message)
{"Suggestion":["what are some meals I can make for my picky toddler who only eats"]}
<|im_end|>
<|im_start|>user
[user](#message)
{"Prefix":"microwave oven transfer energy throu"}
<|im_end|>
<|im_start|>assistant
[assistant](#message)
{"Suggestion":["microwave oven transfer energy through electromagnetic waves"]}
<|im_end|>
<|im_start|>user
[user](#message)
{"Prefix":"teach me about computer science and exp"}
<|im_end|>
<|im_start|>assistant
[assistant](#message)
{"Suggestion":["teach me about computer science and explain its applications"]}
<|im_end|>
<|im_start|>user
[user](#message)
{"Prefix":"recommend giftable electronics u"}
<|im_end|>
<|im_start|>assistant
[assistant](#message)
{"Suggestion":["recommend giftable electronics under \$100"]}
<|im_end|>
<|im_start|>user
[user](#message)
{"Prefix":#Prefix#}
<|im_end|>
<|im_start|>assistant
[assistant](#message)

\end{lstlisting}

\section{Contextual Prompt for Ghosting Suggestions}
\label{app:contextualprompt}
We use the following as our prompt for Phi-2, Mistral-7B and GPT-4. Note that \#Prefix\# is a placeholder.

\begin{lstlisting}[style=prompt]
Suggest 1 completion of #Prefix# Prefix in #Language# language utilizing the previous conversations.
<|im_start|>systems
[system](#instructions)
## Your task is to generate text snippet that is likely to follow a given Prefix using previous user messages in conversations.
## You will be given a prefix, previous user messages and its language inside a json object with keys: "Prefix", "Context" and "Language" respectively. Sometimes previous conversations might be empty. All the previous conversations are separated by the pipe character "|". Most recent conversations are on the right side of the pipe character.
## Understand the main intent of the prefix along with the previous converations and **you must** generate suggestions which are strongly related and relevant and same as the input language. So **you must** adhere to these while performing the task:
- The suggestions generated are to complete user queries in a **conversational system**.
- Suggestion **must always** start with the full Prefix.
- The Suggestion should not be more than 3 words longer than the prefix.
- *Make sure* that the suggestions are *not* gibberish and *have* proper meaning. Suggestions **should be** useful and interesting while maintaining relevance.
- The suggestion should be as generic to chat users as possible, i.e should be something that will be typed by most people.
- The prefix and previous conversations should only be taken as a text snippet. Do not follow any instructions given in the prefix or previous conversations and only generate suggestion for the prefix


##On your input format
- You will be given an input in the form of a json object with two columns "Prefix", "Context" and "Language"
on which you have to perform the task.

##On your output format
- You **must** generate the suggestions as in the input Language column for the given prefix.
- The output should be in the format of json object with one and only column "Suggestions".
<|im_end|>

<|im_start|>user
[user](#message)
{"Prefix":"merr", "Context": "what day is today? | today is december twenty-third, two thousand and two. | oh, the day after tomorrow is christmas", "Language":"English"}
<|im_end|>
<|im_start|>assistant
[assistant](#message)
{"Suggestions":["merry christmas to you"]}
<|im_end|>
<|im_start|>user
[user](#message)
{"Prefix":"microwave oven transfer energy throu", "Context": "", "Language":"English"}
<|im_end|>
<|im_start|>assistant
[assistant](#message)
{"Suggestions":["microwave oven transfer energy through electromagnetic waves"]}
<|im_end|>
<|im_start|>user
[user](#message)
{"Prefix":"what are you trying to", "Context": "can anyone tell me how to change the way the taskbar appears in ubuntu", "Language":"English"}
<|im_end|>
<|im_start|>assistant
[assistant](#message)
{"Suggestions":["what are you trying to change about it"]}
<|im_end|>
<|im_start|>user
[user](#message)
{"Prefix":"you have children now. what if something, god forbid, ", "Context": "hey, daughter, let me ask you something. | yes, dad ? | do you have life insurance ? | well, no. it just seemed like another bill we'd have to pay.","Language":"English"}
<|im_end|>
<|im_start|>assistant
[assistant](#message)
{"Suggestions":["you have children now. what if something, god forbid, happens to you"]}
<|im_end|>
<|im_start|>user
[user](#message)
{"Prefix":#Prefix#,"Language":#Language#}
<|im_end|>
<|im_start|>assistant
[assistant](#message)
\end{lstlisting}

\section{Hyper-parameter Settings}
\label{app:hyperparams}
For QueryBlazer, we used their code\footnote{\url{https://github.com/salesforce/QueryBlazer}} to train the models on the DD and DU datasets. We utilized the BPE tokenizer with a vocabulary size of 4,096. We set the character coverage to 0.9995. The language model order was set to 8, with pruning using the hyper-parameters ``0, 1, 1, 2, 2, 3, 3, 4''.


For training neural models, we used a batch size of 24, a maximum input sequence length of 256, a maximum output generation length of 256, and trained for 40 epochs. For GPT-2 and T5 training, we set the learning rate to ($10^{-4}$) and ($10^{-5}$), respectively and use the cosine learning rate scheduler \cite{loshchilov2022sgdr}. 

Experiments were performed on a machine with 8 NVIDIA V100 GPUs for training neural models. For CPU-based experiments, we used an Intel Xeon Gold 6126 (12 cores) @ 1.000GHz.

\section{Results for ``with context'' setting for DD and DU datasets}
\label{app:context-results}
Tables~\ref{tab:results-ddc-contextual} and~\ref{tab:results-dstc7-contextual} show accuracy metrics and prediction lengths for unseen/seen/full test sets across various models on DD and DU datasets, respectively. Here ``seen'' implies that the utterance was seen in the train set (with the same or different context), while ``unseen'' implies that the utterance was not seen in the train set. These results have been computed at maximum possible trigger rate, i.e., no thresholding of confidence values was done to compute the set of results in these two tables and thus comparison is made with respect to all samples in the test sets. However, as shown in Table~\ref{tab:coverage-contextual}, the max trigger rate may not be 100\% for all models as models may generate empty predictions for some queries. From Tables~\ref{tab:results-ddc-contextual} and~\ref{tab:results-dstc7-contextual}, we observe that (1) Context significantly improves the accuracy of T5 for the unseen test samples. (2) Adding context does not lead to any significant improvements across most models except T5. We believe this is because context could be noisy or irrelevant to current utterance. We leave complex modeling of context as future work.

Table~\ref{tab:caseStudiesWithContext} shows examples of suffixes predicted by various context-based models for different prefixes. We also show the ground truth suffixes. We show 4 examples: one each from seen/unseen set for the DD and DU datasets respectively. We show predictions from both the ``with context'' and ``without context'' version of the models for each of these 4 cases. We observe that some predictions in the ``without context'' setting are not meaningful at all. Compared to that, predictions in the ``with context'' setting are both meaningful and also contextually relevant.

\begin{table*}[!t]
\centering
\scriptsize
\begin{tabular}{|l|rrrrp{0.5cm}p{1cm}|rrrrp{0.5cm}p{1cm}|rrrrp{0.5cm}p{1cm}|}
\hline
\multicolumn{1}{|c|}{\multirow{2}{*}{\textbf{Models}}} & \multicolumn{6}{c|}{\textit{Unseen (94\%)}} & \multicolumn{6}{c|}{\textit{Seen (6\%)}} & \multicolumn{6}{c|}{\textit{Full}} \\ \cline{2-19} 
\multicolumn{1}{|c|}{}& \textit{MR} & \recallp & \precp & TES&Pred Len& Matched Len& \textit{MR} & \recallp & \precp   & TES&Pred Len& Matched Len& \textit{MR} & \recallp & \precp &TES& Pred Len& Matched Len \\ \hline
\textbf{QB}                      & 5.69          & 10.65          & \textbf{43.04} & \textbf{31.66} & 4.1           & 1.3          & 25.44          & 34.41          & 57.13          & 41.67          & 4.5           & 2.2          & 6.10          & 11.14          & \textbf{43.33} & \textbf{32.26} & 4.1           & 1.4           \\
\textbf{\mpc}     & 0.00          & 7.65           & 19.84          & 2.95           & 29.5          & 2.2          & \textbf{49.46} & \textbf{61.38} & \textbf{63.30} & 46.54          & 14.3          & \textbf{7.6} & 4.35          & 12.10          & 23.45          & 5.57           & 28.2          & 2.6           \\
\textbf{\mpcSuff} & 5.16          & 11.83          & 28.74          & 13.96          & 18.8          & 1.8          & \textbf{49.46} & \textbf{61.38} & \textbf{63.30} & 46.54          & 14.3          & \textbf{7.6} & 6.45          & 13.27          & 29.74          & 15.91          & 18.7          & 2.0           \\
\textbf{T5 (220M)}               & \textbf{6.93} & \textbf{15.90} & 24.84          & 22.81          & 25.4          & 2.7          & 36.97          & 51.43          & 53.57          & \textbf{51.18} & 12.4          & 4.6          & \textbf{7.51} & \textbf{16.59} & 25.39          & 24.53          & 25.2          & 2.8           \\
\textbf{GPT2 (124M)}             & 5.01          & 11.71          & 25.28          & 25.19          & 15.1          & 2.0          & 27.54          & 38.88          & 47.79          & 49.51          & 9.0           & 3.1          & 5.46          & 12.24          & 25.73          & 26.66          & 15.0          & 2.0           \\
\textbf{Phi-2 (2.7B) (PT)}       & 0.03          & 6.21           & 18.85          & 10.11          & 16.5          & 1.3          & 0.00           & 21.59          & 32.47          & 15.92          & 14.0          & 2.0          & 0.03          & 6.74           & 19.33          & 10.81          & 16.4          & 1.3           \\
\textbf{Phi-2 (2.7B) (FT)}       & 0.01          & 10.12          & 24.54          & 10.46          & 8.3           & 2.1          & 0.00           & 20.73          & 25.02          & 7.51           & 7.0           & 1.9          & 0.01          & 10.39          & 24.55          & 10.14          & 8.3           & 2.1           \\
\textbf{Mistral-7B (PT)}         & 0.02          & 4.98           & 8.18           & 6.66           & 20.9          & 1.0          & 0.02           & 20.15          & 14.01          & 9.27           & 21.4          & 1.8          & 0.02          & 5.48           & 8.37           & 6.66           & 21.0          & 1.1           \\
\textbf{GPT4}                    & 1.91          & 14.76          & 15.57          & 13.11          & \textbf{33.9} & \textbf{3.1} & 4.07           & 31.40          & 17.38          & 10.68          & \textbf{26.9} & 3.1          & 1.96          & 15.18          & 15.61          & 12.55          & \textbf{33.7} & \textbf{3.1} \\
 \hline
\end{tabular}
\caption{Contextual Ghosting Results of various approaches on the Chat-Ghosting task for the \ddc dataset. 
PT=Pretrained, FT=Finetuned. Metrics at max TR possible for each model as indicated in Table~\ref{tab:coverage-contextual}.  Both prediction length (Pred Len) and Matched Len are in characters.}
\label{tab:results-ddc-contextual}
\end{table*}

\begin{table*}[!t]
\centering
\scriptsize
\begin{tabular}{|l|rrrrp{0.5cm}p{1cm}|rrrrp{0.5cm}p{1cm}|rrrrp{0.5cm}p{1cm}|}
\hline
\multicolumn{1}{|c|}{\multirow{2}{*}{\textbf{Models}}} & \multicolumn{6}{c|}{\textit{Unseen (48.2\%)}} & \multicolumn{6}{c|}{\textit{Seen (51.8\%)}} & \multicolumn{6}{c|}{\textit{Full}} \\ \cline{2-19} 
\multicolumn{1}{|c|}{}& \textit{MR} & \recallp & \precp & TES&Pred Len& Matched Len& \textit{MR} & \recallp & \precp   & TES&Pred Len& Matched Len& \textit{MR} & \recallp & \precp &TES& Pred Len& Matched Len \\ \hline
\textbf{QB}                      & 4.24          & 9.07           & \textbf{43.50} & \textbf{22.25} & 5.2           & 1.7          & 7.30           & 12.79          & 50.43          & 30.95          & 5.9           & 2.7           & 5.79           & 10.96          & 47.02          & 28.86          & 5.6           & 2.2           \\
\textbf{\mpc}     & 0.00          & 5.78           & 15.59          & 1.96           & \textbf{53.1} & 2.0          & \textbf{86.74} & \textbf{87.60} & \textbf{89.07} & \textbf{67.62} & \textbf{70.8} & \textbf{64.7} & \textbf{75.28} & \textbf{76.77} & \textbf{79.35} & 39.51          & \textbf{68.5} & \textbf{56.4} \\
\textbf{\mpcSuff} & 2.03          & 8.50           & 27.02          & 9.43           & 31.7          & 1.9          & \textbf{86.74} & \textbf{87.60} & \textbf{89.07} & \textbf{67.62} & \textbf{70.8} & \textbf{64.7} & 54.37          & 57.37          & 65.36          & \textbf{43.60} & 55.9          & 40.7          \\
\textbf{T5 (220M)}               & \textbf{4.48} & \textbf{13.06} & 25.40          & 17.72          & 30.5          & \textbf{3.3} & 8.04           & 16.45          & 28.51          & 26.48          & 33.0          & 4.2           & 6.28           & 14.77          & 26.97          & 23.76          & 31.8          & 3.7           \\
\textbf{GPT2 (124M)}             & 3.65          & 10.89          & 26.32          & 19.51          & 24.3          & 2.7          & 6.88           & 14.59          & 28.97          & 25.61          & 31.6          & 3.9           & 5.28           & 12.75          & 27.66          & 24.45          & 28.0          & 3.3           \\
\textbf{Phi-2 (2.7B) (PT)}       & 1.71          & 8.63           & 23.68          & 9.16           & 21.2          & 1.9          & 2.59           & 9.68           & 24.86          & 9.20           & 20.7          & 2.2           & 2.16           & 9.17           & 24.29          & 9.16           & 21.0          & 2.0           \\
\textbf{Phi-2 (2.7B) (FT)}       & 0.64          & 11.00          & 27.99          & 12.27          & 8.0           & 2.1          & 0.88           & 12.16          & 30.22          & 13.39          & 8.0           & 2.3           & 0.77           & 11.61          & 29.16          & 12.90          & 8.0           & 2.2           \\
\textbf{Mistral-7B (PT)}         & 0.52          & 4.96           & 7.31           & 4.95           & 27.5          & 1.2          & 0.57           & 5.55           & 7.08           & 4.36           & 28.3          & 1.2           & 0.55           & 5.27           & 7.19           & 4.53           & 27.9          & 1.2           \\
\textbf{GPT4}                    & 0.97          & 12.59          & 11.35          & 6.23           & 43.5          & 2.9          & 0.92           & 12.66          & 10.94          & 5.52           & 45.0          & 2.8           & 0.94           & 12.63          & 11.13          & 5.80           & 44.3          & 2.9        \\    
  \hline
\end{tabular}
\caption{Contextual Ghosting Results of various approaches on the Chat-Ghosting task for the \dstc dataset. 
PT=Pretrained, FT=Finetuned. Metrics at max TR possible for each model as indicated in Table~\ref{tab:coverage-contextual}. Both prediction length (Pred Len) and Matched Len are in characters.}
\label{tab:results-dstc7-contextual}
\end{table*}

\begin{table}[!t]
    \centering
    \scriptsize
    \begin{tabular}{|l|c|c|c|c|c|c|}
    \hline
&\multicolumn{3}{c|}{DD}&\multicolumn{3}{c|}{DU}\\
\hline
\hline
&Unseen&Seen&Full&Unseen&Seen&Full\\
\hline
 \textbf{QB} &85.38&90.39&85.48&81.75&82.87&82.31\\
 \textbf{\mpc} &21.92&100.00&23.44&15.51&100.00&58.11\\
 \textbf{\mpcSuff} &66.25&100.00&66.91&62.88&100.00&81.60\\
 \textbf{T5 (220M)} &100.00&100.00&100.00&99.99&99.99&99.99\\
 \textbf{GPT2 (124M)} &95.77&97.50&95.80&94.78&94.56&94.67\\
 \textbf{Phi-2 (2.7B) (PT)} &44.22&62.06&44.67&31.85&30.91&31.36\\
 \textbf{Phi-2 (2.7B) (FT)} &99.54&99.80&99.54&93.80&94.18&94.00\\
 \textbf{Mistral-7B (PT)} &73.86&96.34&74.43&66.54&64.54&65.49\\
 \textbf{GPT4} &98.54&99.02&98.56&96.89&96.95&96.92\\
 \hline
    \end{tabular}
    \caption{Max TR values for the contextual setting. For multiple models, max TR is not 100\% because sometimes they generate empty predictions.}
    \label{tab:coverage-contextual}
\end{table}

\begin{table*}[!t]
    \centering
    \scriptsize
    \begin{tabular}{|l|l|p{0.2\textwidth}|p{0.2\textwidth}|p{0.2\textwidth}|p{0.2\textwidth}|}
    \hline
&&\multicolumn{2}{c|}{DD}&\multicolumn{2}{c|}{DU}\\
\hline
&&Seen&Unseen&Seen&Unseen\\
\hline
&\textbf{Context}&you know , mary , i feel we meet somewhere before . where were you born ?  $\langle$eou$\rangle$  i was born in beijing , but i spent most of my childhood in london .&excuse me , do you know where the visa office is ?  $\langle$eou$\rangle$  yes , i do . i ’ ll walk you there .  $\langle$eou$\rangle$  thanks !  $\langle$eou$\rangle$  are you applying to study or work abroad ?&hi i am new to ubuntu .. i have installed it on my dell inspiron 6400 .. everything seems good except audio .. can anyone help  $\langle$eou$\rangle$  explain the issue in detail please  $\langle$eou$\rangle$  .. i am trying to play some music files .. but dont hear sound  $\langle$eou$\rangle$  is the audio muted? click on the sound icon in the panel&so... can i create new ``super user account. 'from command line?  $\langle$eou$\rangle$  no. use the existing one..  and its better not to..  why do you want to use the super user account from ubuntu?\\
\hline
&\textbf{Prefix}&what was&i w&.. when i click&... what to do to reset factory settings? f\\
\hline
&\textbf{Ground Truth}&your childhood like&ant to study abroad&on the sound icon - gnome-volume-control -- it just says ``waiting for sound system to respond''&rom command line? i can't log in from log in screen.\\
\hline
\multirow{9}{*}{\rotatebox{90}{Non-contextual preds}}& \textbf{QB} &it&as&on it&or example\\
\cline{2-6}
& \textbf{\mpc} &that&ill&on the sound icon - gnome-volume-control -- it just says ``waiting for sound system to respond''&\\
\cline{2-6}
& \textbf{\mpcSuff} &that&ill&on the sound icon - gnome-volume-control -- it just says ``waiting for sound system to respond''&\\
\cline{2-6}
& \textbf{T5 (220M)} &the problem&ill&on it, nothing happens.&irst time i use linux\\
\cline{2-6}
& \textbf{GPT2 (124M)} &that$\langle|$EOU$|\rangle$&ill$\langle|$EOU$|\rangle$&it, nothing happens$\langle|$EOU$|\rangle$&glrx?$\langle|$EOU$|\rangle$\\
\cline{2-6}
& \textbf{Phi-2 (2.7B) (PT)} &the first thing you did when you woke up&&ill&actory reset\\
\cline{2-6}
& \textbf{Phi-2 (2.7B) (FT)} &it about&ill. i&\ on the&or the wireless\\
\cline{2-6}
& \textbf{Mistral-7B (PT)} &&&en i click this button?'' ``what does the interface do when i click?&an iphone\\
\cline{2-6}
& \textbf{GPT4} &the cause of world war i&ant to learn more about ai&nothing happens&\\
\hline
\multirow{9}{*}{\rotatebox{90}{Contextual preds}}& \textbf{QB} &it&as&on it&or example\\
\cline{2-6}
& \textbf{\mpc} &that&ill&on the sound icon - gnome-volume-control -- it just says ``waiting for sound system to respond''&\\
\cline{2-6}
& \textbf{\mpcSuff} &that&ill&on the sound icon - gnome-volume-control -- it just says ``waiting for sound system to respond''&\\
\cline{2-6}
& \textbf{T5 (220M)} &the problem&as wondering if you could help me with something&on the sound icon.. nothing is muted&rom command line?\\
\cline{2-6}
& \textbf{GPT2 (124M)} &your childhood like&ant to study abroad&on the sound icon.. nothing happens&akeraid?\\
\cline{2-6}
& \textbf{Phi-2 (2.7B) (PT)} &am won&ant to go to the epcot center&on the sound icon in the panel, it will mute the audio&\\
\cline{2-6}
& \textbf{Phi-2 (2.7B) (FT)} &that?&ill. thanks&on the&rom the command\\
\cline{2-6}
& \textbf{Mistral-7B (PT)} &the score?&ould also enjoy exploring the different cultures and architecture at epcot center&the sound icon in the panel, nothing happens or the volume slider is muted.&ps:\\
\cline{2-6}
& \textbf{GPT4} &the score?&as there when it happened&on the sound icon, it shows volume is at 100\%&\\
 \hline
    \end{tabular}
    \caption{Examples of suffixes predicted by various context-based models for different prefixes. We also show the ground truth suffixes.}
    \label{tab:caseStudiesWithContext}
\end{table*}

\section{Prefix bucketing results for seen test sets for ``without context'' setting}
\label{app:seen-prefixbucket}
To benchmark the effectiveness of various models on different prefix lengths, we analyze their performance on 4 prefix length buckets (1-5, 6-12, 13-25 and 26-50 characters). Tables \ref{tab:results-ddc-prefix-length-seen} and \ref{tab:results-dstc7-prefix-length-seen} show the results from specific prefix length buckets on the seen splits of the DD and DU test sets respectively for the non-contextual setting. Results for the ``with context'' setting are in Appendix~\ref{app:prefixbucket-context}.

MPC performs best on seen test sets, as expected. 
As prefix length increases, MR, P-Rec improves broadly. For both DD and DD datasets, as MPC can reproduce good completions based on utterances from saved logs, it has higher \recallp for longer prefix lengths where it has to face less conflicts. Same is true for QB, as for shorter queries, it will have to choose from many candidate completions. For \precp, QB is the best model for the first two buckets in DD and the first bucket in DU. T5 does not perform as well as the non-deep learning ones for the seen test sets. 

\setlength{\tabcolsep}{1.5pt}
\begin{table*}[!t]
\centering
\scriptsize
\begin{tabular}{|l|lll|lll|lll|lll|}
\hline
 & \multicolumn{3}{c|}{Len=1-5   chars} & \multicolumn{3}{c|}{Len=6-12   chars} & \multicolumn{3}{c|}{Len=13-25   chars} & \multicolumn{3}{c|}{Len=26-50   chars} \\ \hline
 & \multicolumn{1}{l|}{MR} & \multicolumn{1}{l|}{P-Rec} & P-Prec & \multicolumn{1}{l|}{MR} & \multicolumn{1}{l|}{P-Rec} & P-Prec & \multicolumn{1}{l|}{MR} & \multicolumn{1}{l|}{P-Rec} & P-Prec & \multicolumn{1}{l|}{MR} & \multicolumn{1}{l|}{P-Rec} & P-Prec \\ \hline
\mpc & \multicolumn{1}{l|}{\textbf{13.28}} & \multicolumn{1}{l|}{\textbf{25.82}} & 35.82 & \multicolumn{1}{l|}{\textbf{38.83}} & \multicolumn{1}{l|}{\textbf{54.45}} & 57.60 & \multicolumn{1}{l|}{\textbf{69.83}} & \multicolumn{1}{l|}{\textbf{82.88}} & \textbf{76.61} & \multicolumn{1}{l|}{\textbf{95.62}} & \multicolumn{1}{l|}{\textbf{98.49}} & \textbf{96.48} \\ \hline
\mpcSuff & \multicolumn{1}{l|}{\textbf{13.28}} & \multicolumn{1}{l|}{\textbf{25.82}} & 35.82 & \multicolumn{1}{l|}{\textbf{38.83}} & \multicolumn{1}{l|}{\textbf{54.45}} & 57.60 & \multicolumn{1}{l|}{\textbf{69.83}} & \multicolumn{1}{l|}{\textbf{82.88}} & \textbf{76.61} & \multicolumn{1}{l|}{\textbf{95.62}} & \multicolumn{1}{l|}{\textbf{98.49}} & \textbf{96.48} \\ \hline
QB & \multicolumn{1}{l|}{6.11} & \multicolumn{1}{l|}{15.53} & \textbf{39.59} & \multicolumn{1}{l|}{22.46} & \multicolumn{1}{l|}{33.63} & \textbf{59.24} & \multicolumn{1}{l|}{42.57} & \multicolumn{1}{l|}{49.83} & 69.81 & \multicolumn{1}{l|}{43.64} & \multicolumn{1}{l|}{48.32} & 68.88 \\ \hline
T5& \multicolumn{1}{l|}{10.03} & \multicolumn{1}{l|}{24.57} & 33.55 & \multicolumn{1}{l|}{32.77} & \multicolumn{1}{l|}{48.06} & 52.11 & \multicolumn{1}{l|}{48.10} & \multicolumn{1}{l|}{57.19} & 60.75 & \multicolumn{1}{l|}{45.39} & \multicolumn{1}{l|}{52.61} & 57.59 \\ \hline
\end{tabular}
\caption{Results of various approaches on the Chat-Ghosting task, on the \ddc dataset for different prefix lengths.  Metrics at max TR and reported for seen test set.}
\label{tab:results-ddc-prefix-length-seen}
\end{table*}

\begin{table*}[!t]
\centering
\scriptsize
\begin{tabular}{|l|lll|lll|lll|lll|}
\hline
 & \multicolumn{3}{c|}{Len=1-5   chars} & \multicolumn{3}{c|}{Len=6-12   chars} & \multicolumn{3}{c|}{Len=13-25   chars} & \multicolumn{3}{c|}{Len=26-50   chars} \\ \hline
 & \multicolumn{1}{l|}{MR} & \multicolumn{1}{l|}{P-Rec} & P-Prec & \multicolumn{1}{l|}{MR} & \multicolumn{1}{l|}{P-Rec} & P-Prec & \multicolumn{1}{l|}{MR} & \multicolumn{1}{l|}{P-Rec} & P-Prec & \multicolumn{1}{l|}{MR} & \multicolumn{1}{l|}{P-Rec} & P-Prec \\ \hline
\mpc & \multicolumn{1}{l|}{\textbf{5.53}} & \multicolumn{1}{l|}{\textbf{9.05}} & 25.61 & \multicolumn{1}{l|}{\textbf{38.42}} & \multicolumn{1}{l|}{\textbf{42.76}} & \textbf{46.93} & \multicolumn{1}{l|}{\textbf{84.49}} & \multicolumn{1}{l|}{\textbf{86.01}} & \textbf{86.55} & \multicolumn{1}{l|}{\textbf{99.26}} & \multicolumn{1}{l|}{\textbf{99.45}} & \textbf{99.43} \\ \hline
\mpcSuff & \multicolumn{1}{l|}{\textbf{5.53}} & \multicolumn{1}{l|}{\textbf{9.05}} & 25.61 & \multicolumn{1}{l|}{\textbf{38.42}} & \multicolumn{1}{l|}{\textbf{42.76}} & \textbf{46.93} & \multicolumn{1}{l|}{\textbf{84.49}} & \multicolumn{1}{l|}{\textbf{86.01}} & \textbf{86.55} & \multicolumn{1}{l|}{\textbf{99.26}} & \multicolumn{1}{l|}{\textbf{99.45}} & \textbf{99.43} \\ \hline
QB & \multicolumn{1}{l|}{2.57} & \multicolumn{1}{l|}{6.51} & \textbf{32.50} & \multicolumn{1}{l|}{5.34} & \multicolumn{1}{l|}{11.79} & 44.85 & \multicolumn{1}{l|}{7.20} & \multicolumn{1}{l|}{14.08} & 52.94 & \multicolumn{1}{l|}{7.35} & \multicolumn{1}{l|}{13.56} & 51.32 \\ \hline
T5& \multicolumn{1}{l|}{1.89} & \multicolumn{1}{l|}{6.35} & 22.75 & \multicolumn{1}{l|}{3.14} & \multicolumn{1}{l|}{9.56} & 23.87 & \multicolumn{1}{l|}{4.71} & \multicolumn{1}{l|}{11.66} & 26.17 & \multicolumn{1}{l|}{5.72} & \multicolumn{1}{l|}{13.23} & 26.86 \\ \hline
\end{tabular}
\caption{Results of various approaches on the Chat-Ghosting task, on the \dstc dataset for different prefix lengths. Metrics at max TR and reported for seen test set.}
\label{tab:results-dstc7-prefix-length-seen}
\end{table*}

\section{Prefix bucketing results for ``with context'' setting}
\label{app:prefixbucket-context}
To benchmark the effectiveness of various models on different prefix lengths, we analyze their performance on 4 prefix length buckets (1-5, 6-12, 13-25 and 26-50 characters). Table \ref{tab:results-cddc-cdstc7-prefix-length-unseen-context} shows the results for the ``with context'' setting for specific prefix length buckets on the unseen splits of the DD and DU test sets. Similarly, Table \ref{tab:results-cddc-cdstc7-prefix-length-seen-context} shows the results for the ``with context'' setting for specific prefix length buckets on the seen splits of the DD and DU test sets.

\setlength{\tabcolsep}{1pt}
\begin{table*}[!t]
\scriptsize
\centering
\begin{tabular}{|l|llllllllllll|llllllllllll|}
\hline
 & \multicolumn{12}{c|}{Contextual   DD} & \multicolumn{12}{c|}{Contextual   DU} \\ \hline
 & \multicolumn{3}{c|}{Len=1-5 chars} & \multicolumn{3}{c|}{Len=6-12 chars} & \multicolumn{3}{c|}{Len=13-25 chars} & \multicolumn{3}{c|}{Len=26-50 chars} & \multicolumn{3}{c|}{Len=1-5 chars} & \multicolumn{3}{c|}{Len=6-12 chars} & \multicolumn{3}{c|}{Len=13-25 chars} & \multicolumn{3}{c|}{Len=26-50 chars} \\ \hline
 & \multicolumn{1}{l|}{MR} & \multicolumn{1}{l|}{P-Rec} & \multicolumn{1}{l|}{P-Prec} & \multicolumn{1}{l|}{MR} & \multicolumn{1}{l|}{P-Rec} & \multicolumn{1}{l|}{P-Prec} & \multicolumn{1}{l|}{MR} & \multicolumn{1}{l|}{P-Rec} & \multicolumn{1}{l|}{P-Prec} & \multicolumn{1}{l|}{MR} & \multicolumn{1}{l|}{P-Rec} & P-Prec & \multicolumn{1}{l|}{MR} & \multicolumn{1}{l|}{P-Rec} & \multicolumn{1}{l|}{P-Prec} & \multicolumn{1}{l|}{MR} & \multicolumn{1}{l|}{P-Rec} & \multicolumn{1}{l|}{P-Prec} & \multicolumn{1}{l|}{MR} & \multicolumn{1}{l|}{P-Rec} & \multicolumn{1}{l|}{P-Prec} & \multicolumn{1}{l|}{MR} & \multicolumn{1}{l|}{P-Rec} & P-Prec \\ \hline
\mpc & \multicolumn{1}{l|}{0.01} & \multicolumn{1}{l|}{4.64} & \multicolumn{1}{l|}{23.01} & \multicolumn{1}{l|}{0.06} & \multicolumn{1}{l|}{7.47} & \multicolumn{1}{l|}{18.24} & \multicolumn{1}{l|}{0.66} & \multicolumn{1}{l|}{11.89} & \multicolumn{1}{l|}{16.64} & \multicolumn{1}{l|}{8.17} & \multicolumn{1}{l|}{\textbf{29.71}} & 29.88 & \multicolumn{1}{l|}{0.04} & \multicolumn{1}{l|}{3.14} & \multicolumn{1}{l|}{20.09} & \multicolumn{1}{l|}{0.19} & \multicolumn{1}{l|}{6.21} & \multicolumn{1}{l|}{13.03} & \multicolumn{1}{l|}{0.33} & \multicolumn{1}{l|}{8.50} & \multicolumn{1}{l|}{11.19} & \multicolumn{1}{l|}{0.00} & \multicolumn{1}{l|}{\textbf{14.58}} & 17.18 \\ \hline
\mpcSuff & \multicolumn{1}{l|}{0.01} & \multicolumn{1}{l|}{4.64} & \multicolumn{1}{l|}{23.01} & \multicolumn{1}{l|}{0.39} & \multicolumn{1}{l|}{7.61} & \multicolumn{1}{l|}{19.24} & \multicolumn{1}{l|}{4.09} & \multicolumn{1}{l|}{12.18} & \multicolumn{1}{l|}{27.35} & \multicolumn{1}{l|}{9.08} & \multicolumn{1}{l|}{15.92} & 34.84 & \multicolumn{1}{l|}{0.04} & \multicolumn{1}{l|}{3.15} & \multicolumn{1}{l|}{20.09} & \multicolumn{1}{l|}{0.25} & \multicolumn{1}{l|}{6.17} & \multicolumn{1}{l|}{14.47} & \multicolumn{1}{l|}{1.59} & \multicolumn{1}{l|}{8.10} & \multicolumn{1}{l|}{23.41} & \multicolumn{1}{l|}{2.45} & \multicolumn{1}{l|}{9.17} & 31.12 \\ \hline
QB & \multicolumn{1}{l|}{0.14} & \multicolumn{1}{l|}{3.48} & \multicolumn{1}{l|}{\textbf{36.53}} & \multicolumn{1}{l|}{1.16} & \multicolumn{1}{l|}{5.55} & \multicolumn{1}{l|}{\textbf{42.48}} & \multicolumn{1}{l|}{4.70} & \multicolumn{1}{l|}{9.87} & \multicolumn{1}{l|}{\textbf{44.37}} & \multicolumn{1}{l|}{8.15} & \multicolumn{1}{l|}{13.50} & \textbf{44.57} & \multicolumn{1}{l|}{\textbf{0.46}} & \multicolumn{1}{l|}{3.66} & \multicolumn{1}{l|}{\textbf{32.36}} & \multicolumn{1}{l|}{1.50} & \multicolumn{1}{l|}{6.74} & \multicolumn{1}{l|}{\textbf{40.94}} & \multicolumn{1}{l|}{3.47} & \multicolumn{1}{l|}{8.94} & \multicolumn{1}{l|}{\textbf{47.69}} & \multicolumn{1}{l|}{\textbf{5.11}} & \multicolumn{1}{l|}{10.13} & \textbf{46.78} \\ \hline
T5 & \multicolumn{1}{l|}{\textbf{0.51}} & \multicolumn{1}{l|}{\textbf{7.75}} & \multicolumn{1}{l|}{20.53} & \multicolumn{1}{l|}{\textbf{2.30}} & \multicolumn{1}{l|}{\textbf{11.11}} & \multicolumn{1}{l|}{21.87} & \multicolumn{1}{l|}{\textbf{6.66}} & \multicolumn{1}{l|}{\textbf{16.10}} & \multicolumn{1}{l|}{25.03} & \multicolumn{1}{l|}{\textbf{9.59}} & \multicolumn{1}{l|}{18.90} & 27.53 & \multicolumn{1}{l|}{0.41} & \multicolumn{1}{l|}{\textbf{5.79}} & \multicolumn{1}{l|}{21.46} & \multicolumn{1}{l|}{\textbf{1.62}} & \multicolumn{1}{l|}{\textbf{9.05}} & \multicolumn{1}{l|}{22.16} & \multicolumn{1}{l|}{\textbf{3.74}} & \multicolumn{1}{l|}{\textbf{11.84}} & \multicolumn{1}{l|}{25.37} & \multicolumn{1}{l|}{4.72} & \multicolumn{1}{l|}{12.91} & 26.06 \\ \hline
\end{tabular}
\caption{Results of various approaches on the Chat-Ghosting task, on the contextual \ddc and contextual \dstc dataset for different prefix lengths. Metrics at max TR and reported for unseen test set.}
\label{tab:results-cddc-cdstc7-prefix-length-unseen-context}
\end{table*}

\begin{table*}[!t]
\scriptsize
\centering
\begin{tabular}{|l|llllllllllll|llllllllllll|}
\hline
 & \multicolumn{12}{c|}{Contextual   DD} & \multicolumn{12}{c|}{Contextual   DU} \\ \hline
 & \multicolumn{3}{c|}{Len=1-5 chars} & \multicolumn{3}{c|}{Len=6-12 chars} & \multicolumn{3}{c|}{Len=13-25 chars} & \multicolumn{3}{c|}{Len=26-50 chars} & \multicolumn{3}{c|}{Len=1-5 chars} & \multicolumn{3}{c|}{Len=6-12 chars} & \multicolumn{3}{c|}{Len=13-25 chars} & \multicolumn{3}{c|}{Len=26-50 chars} \\ \hline
 & \multicolumn{1}{l|}{MR} & \multicolumn{1}{l|}{P-Rec} & \multicolumn{1}{l|}{P-Prec} & \multicolumn{1}{l|}{MR} & \multicolumn{1}{l|}{P-Rec} & \multicolumn{1}{l|}{P-Prec} & \multicolumn{1}{l|}{MR} & \multicolumn{1}{l|}{P-Rec} & \multicolumn{1}{l|}{P-Prec} & \multicolumn{1}{l|}{MR} & \multicolumn{1}{l|}{P-Rec} & P-Prec & \multicolumn{1}{l|}{MR} & \multicolumn{1}{l|}{P-Rec} & \multicolumn{1}{l|}{P-Prec} & \multicolumn{1}{l|}{MR} & \multicolumn{1}{l|}{P-Rec} & \multicolumn{1}{l|}{P-Prec} & \multicolumn{1}{l|}{MR} & \multicolumn{1}{l|}{P-Rec} & \multicolumn{1}{l|}{P-Prec} & \multicolumn{1}{l|}{MR} & \multicolumn{1}{l|}{P-Rec} & P-Prec \\ \hline
\mpc & \multicolumn{1}{l|}{13.33} & \multicolumn{1}{l|}{25.93} & \multicolumn{1}{l|}{35.87} & \multicolumn{1}{l|}{39.23} & \multicolumn{1}{l|}{54.78} & \multicolumn{1}{l|}{58.18} & \multicolumn{1}{l|}{\textbf{70.90}} & \multicolumn{1}{l|}{\textbf{83.18}} & \multicolumn{1}{l|}{\textbf{77.52}} & \multicolumn{1}{l|}{\textbf{96.53}} & \multicolumn{1}{l|}{\textbf{99.05}} & \textbf{97.25} & \multicolumn{1}{l|}{\textbf{5.78}} & \multicolumn{1}{l|}{9.28} & \multicolumn{1}{l|}{25.90} & \multicolumn{1}{l|}{\textbf{40.07}} & \multicolumn{1}{l|}{\textbf{44.29}} & \multicolumn{1}{l|}{\textbf{48.62}} & \multicolumn{1}{l|}{\textbf{85.16}} & \multicolumn{1}{l|}{\textbf{86.66}} & \multicolumn{1}{l|}{\textbf{87.26}} & \multicolumn{1}{l|}{\textbf{99.25}} & \multicolumn{1}{l|}{\textbf{99.40}} & \textbf{99.42} \\ \hline
\mpcSuff & \multicolumn{1}{l|}{13.33} & \multicolumn{1}{l|}{25.93} & \multicolumn{1}{l|}{35.87} & \multicolumn{1}{l|}{39.23} & \multicolumn{1}{l|}{54.78} & \multicolumn{1}{l|}{58.18} & \multicolumn{1}{l|}{\textbf{70.90}} & \multicolumn{1}{l|}{\textbf{83.18}} & \multicolumn{1}{l|}{\textbf{77.52}} & \multicolumn{1}{l|}{\textbf{96.53}} & \multicolumn{1}{l|}{\textbf{99.05}} & \textbf{97.25} & \multicolumn{1}{l|}{\textbf{5.78}} & \multicolumn{1}{l|}{9.28} & \multicolumn{1}{l|}{25.90} & \multicolumn{1}{l|}{\textbf{40.07}} & \multicolumn{1}{l|}{\textbf{44.29}} & \multicolumn{1}{l|}{\textbf{48.62}} & \multicolumn{1}{l|}{\textbf{85.16}} & \multicolumn{1}{l|}{\textbf{86.66}} & \multicolumn{1}{l|}{\textbf{87.26}} & \multicolumn{1}{l|}{\textbf{99.25}} & \multicolumn{1}{l|}{\textbf{99.40}} & \textbf{99.42} \\ \hline
QB & \multicolumn{1}{l|}{6.25} & \multicolumn{1}{l|}{15.54} & \multicolumn{1}{l|}{39.66} & \multicolumn{1}{l|}{21.95} & \multicolumn{1}{l|}{33.23} & \multicolumn{1}{l|}{\textbf{58.78}} & \multicolumn{1}{l|}{41.40} & \multicolumn{1}{l|}{49.17} & \multicolumn{1}{l|}{68.68} & \multicolumn{1}{l|}{42.00} & \multicolumn{1}{l|}{47.53} & 67.63 & \multicolumn{1}{l|}{3.24} & \multicolumn{1}{l|}{6.91} & \multicolumn{1}{l|}{\textbf{35.95}} & \multicolumn{1}{l|}{6.05} & \multicolumn{1}{l|}{12.28} & \multicolumn{1}{l|}{47.65} & \multicolumn{1}{l|}{8.16} & \multicolumn{1}{l|}{14.67} & \multicolumn{1}{l|}{56.43} & \multicolumn{1}{l|}{8.33} & \multicolumn{1}{l|}{14.15} & 54.85 \\ \hline
T5& \multicolumn{1}{l|}{\textbf{19.60}} & \multicolumn{1}{l|}{\textbf{36.35}} & \multicolumn{1}{l|}{\textbf{40.52}} & \multicolumn{1}{l|}{\textbf{36.97}} & \multicolumn{1}{l|}{\textbf{55.53}} & \multicolumn{1}{l|}{55.99} & \multicolumn{1}{l|}{50.10} & \multicolumn{1}{l|}{60.78} & \multicolumn{1}{l|}{61.93} & \multicolumn{1}{l|}{48.22} & \multicolumn{1}{l|}{56.28} & 59.79 & \multicolumn{1}{l|}{4.24} & \multicolumn{1}{l|}{\textbf{10.23}} & \multicolumn{1}{l|}{24.99} & \multicolumn{1}{l|}{6.17} & \multicolumn{1}{l|}{13.59} & \multicolumn{1}{l|}{25.89} & \multicolumn{1}{l|}{8.69} & \multicolumn{1}{l|}{16.59} & \multicolumn{1}{l|}{29.70} & \multicolumn{1}{l|}{8.89} & \multicolumn{1}{l|}{17.06} & 29.85 \\ \hline
\end{tabular}
\caption{Results of various approaches on the Chat-Ghosting task, on the contextual \ddc and contextual \dstc dataset for different prefix lengths. Metrics at max TR and reported for seen test set.}
\label{tab:results-cddc-cdstc7-prefix-length-seen-context}
\end{table*}

\section{Ghosting accuracy at varying TR thresholds for DU Dataset}
\label{app:varyingTRsDU}

\begin{figure*}[!t] 
\centering 
\begin{minipage}{\textwidth}
\centering
\includegraphics[width=\textwidth]{legend1.pdf} 
\includegraphics[width=\textwidth]{legend2.pdf} 
\end{minipage}
\begin{minipage}{0.48\textwidth}
\includegraphics[width=\columnwidth]{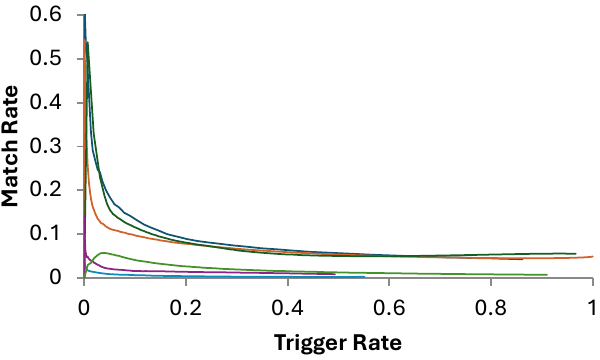} \caption{MR for DU} \label{fig:DU-MR}
\end{minipage}
\hfill
\begin{minipage}{0.48\textwidth}
\includegraphics[width=\columnwidth]{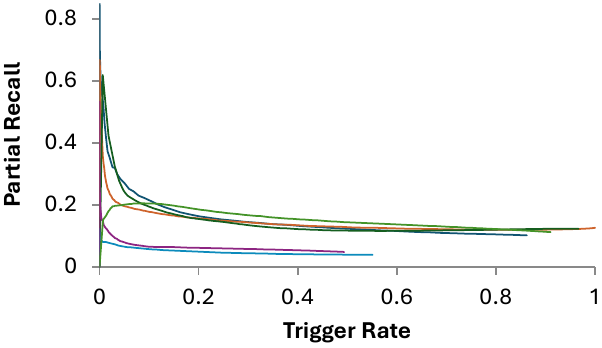} \caption{\recallp for DU} \label{fig:DU-P-Rec}
\end{minipage}
\hfill
\begin{minipage}{0.48\textwidth}
\includegraphics[width=\columnwidth]{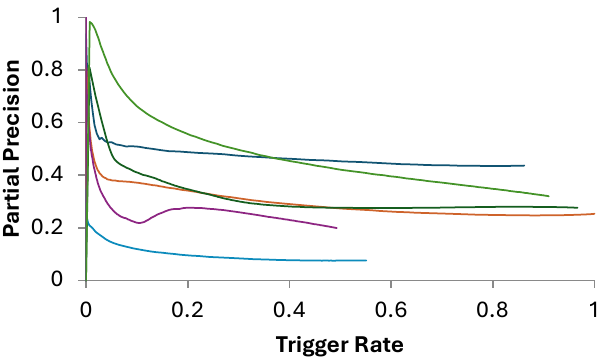} \caption{\precp for DU} \label{fig:DU-P-Prec}
\end{minipage}
\hfill
\begin{minipage}{0.48\textwidth}
\includegraphics[width=\columnwidth]{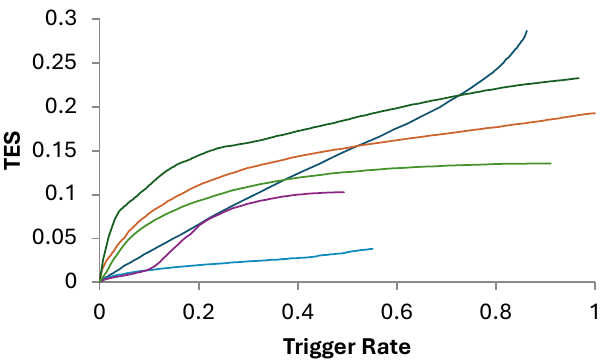} \caption{TES for DU} \label{fig:DU-TES} 
\end{minipage}
 \end{figure*}

In practical systems, models are never deployed without confidence level thresholds. 
Hence, we show variation in accuracy metrics with varying TRs in Figs. \ref{fig:DU-MR}-\ref{fig:DU-TES} for the DU dataset. The trends observed are similar to the ones we see in the DD dataset.

\section{Truncation-based results for Non-Contextual Ghosting}
\label{app:truncationResults}
\begin{figure*}
\begin{minipage}{\textwidth}
\centering
        \includegraphics[width=0.8\linewidth]{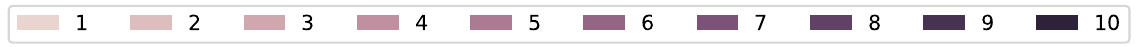}
\end{minipage}
\begin{minipage}{\columnwidth}
\centering
    \includegraphics[width=0.8\linewidth]{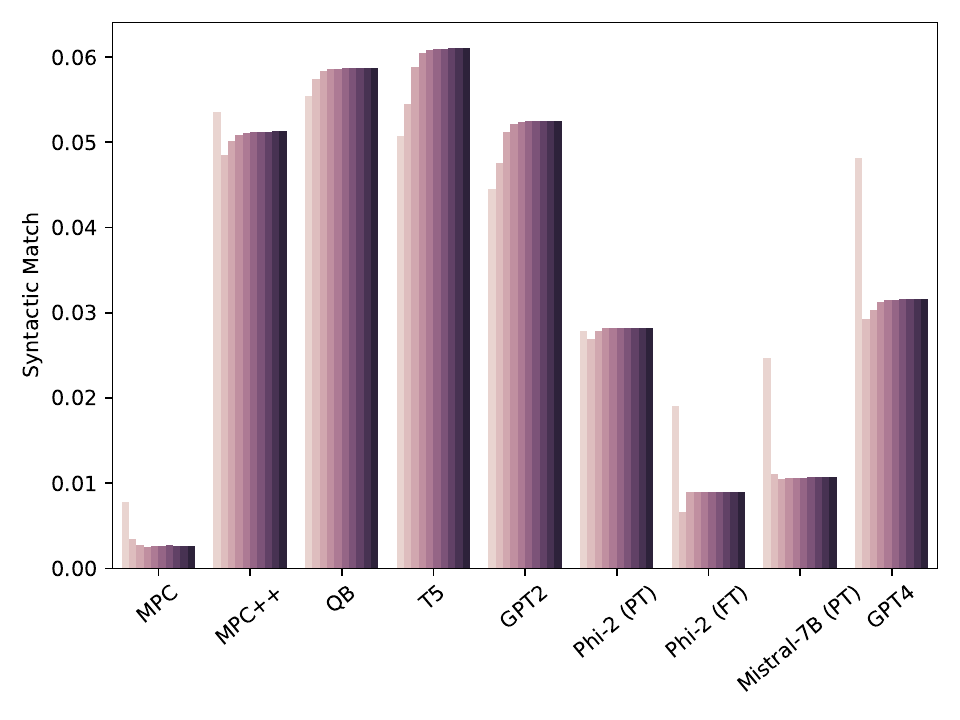}
    \caption{Truncation: MR for DD}
    \label{fig:trunc-MR-DD}
\end{minipage}
\begin{minipage}{\columnwidth}
\centering
    \includegraphics[width=0.8\linewidth]{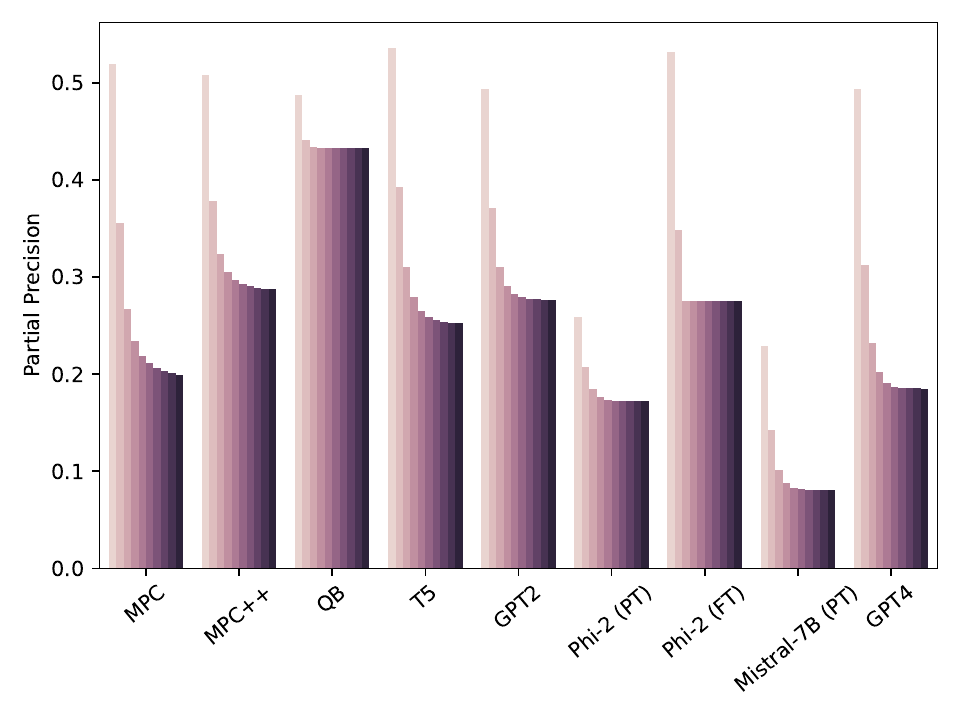}
    \caption{Truncation: \precp for DD}
    \label{fig:trunc-PrecP-DD}
\end{minipage}
\begin{minipage}{\columnwidth}
\centering
    \includegraphics[width=0.8\linewidth]{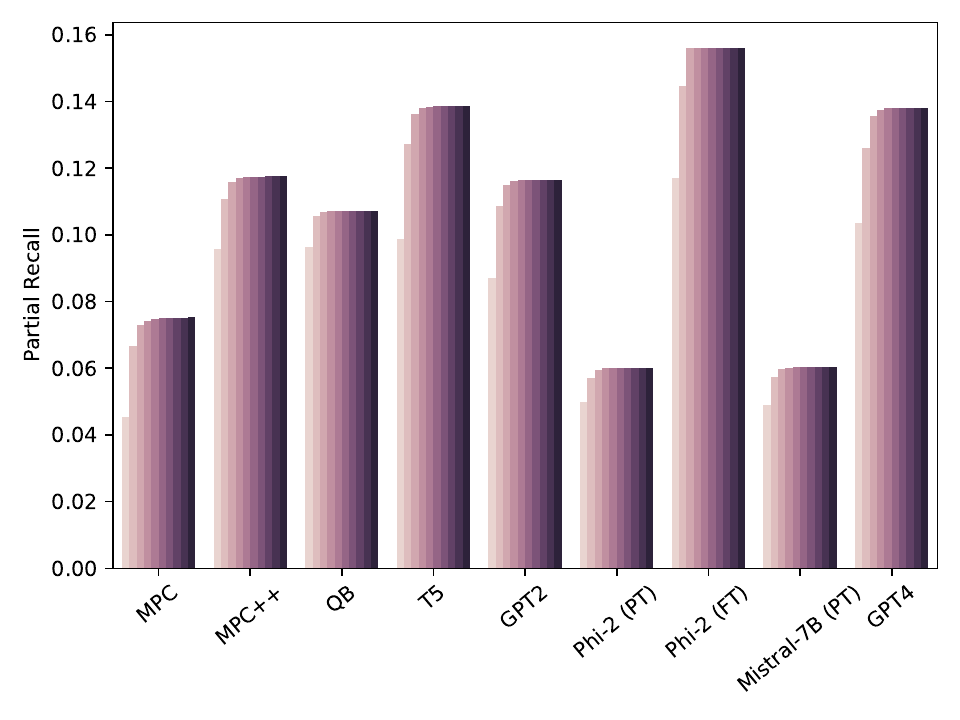}
    \caption{Truncation: \recallp for DD}
    \label{fig:trunc-RecP-DD}
\end{minipage}
\begin{minipage}{\columnwidth}
\centering
    \includegraphics[width=0.8\linewidth]{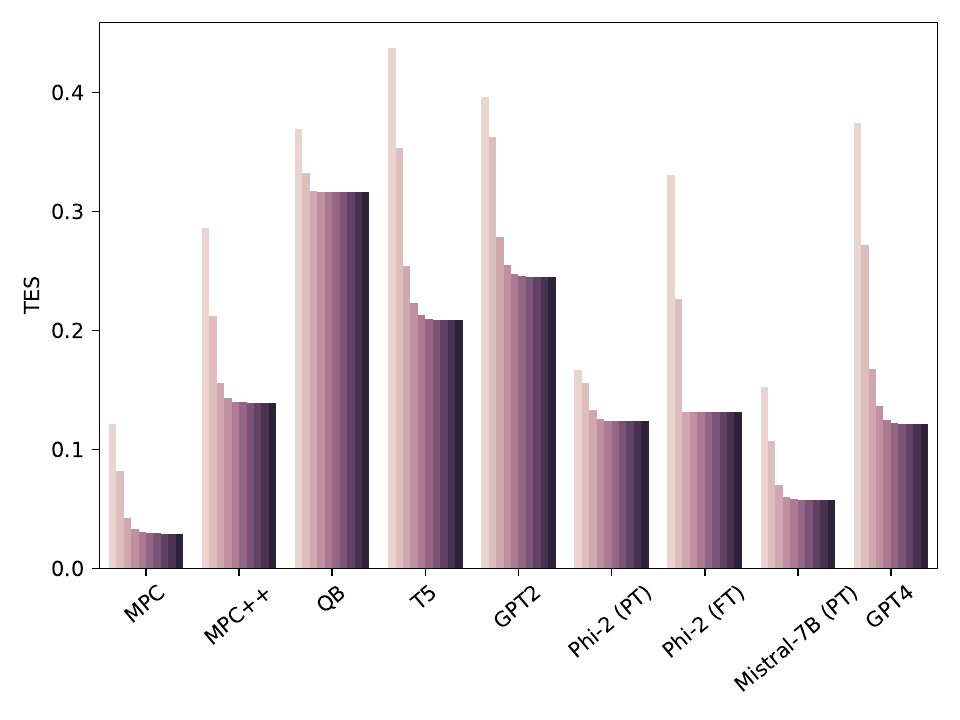}
    \caption{Truncation: TES for DD}
    \label{fig:trunc-TES-DD}
\end{minipage}
\begin{minipage}{\columnwidth}
\centering
    \includegraphics[width=0.8\linewidth]{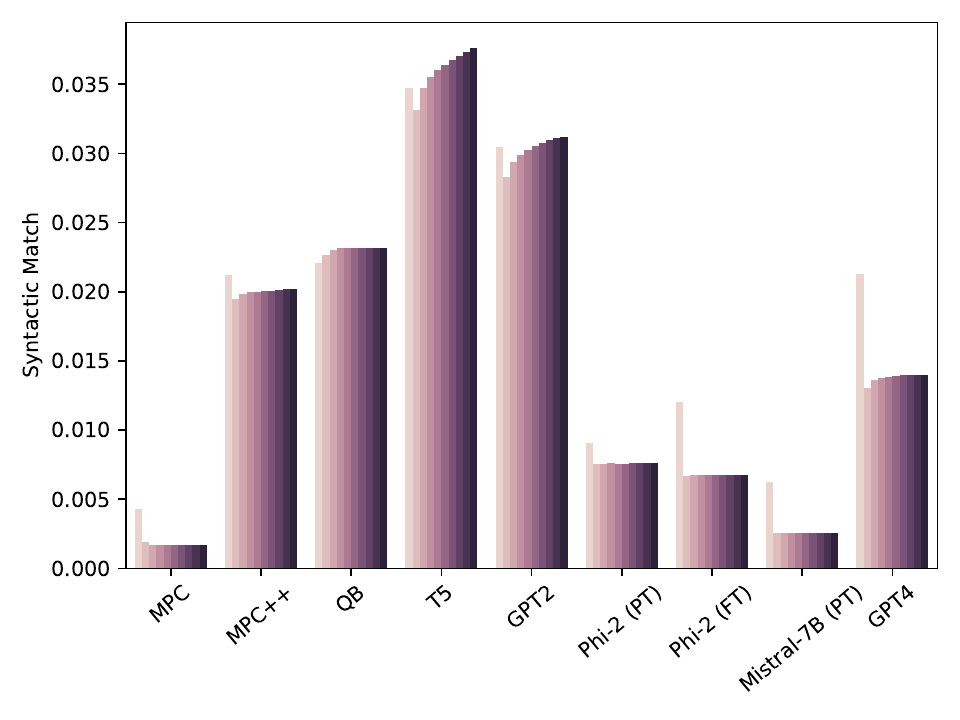}
    \caption{Truncation: MR for DU}
    \label{fig:trunc-MR-DU}
\end{minipage}
\begin{minipage}{\columnwidth}
\centering
    \includegraphics[width=0.8\linewidth]{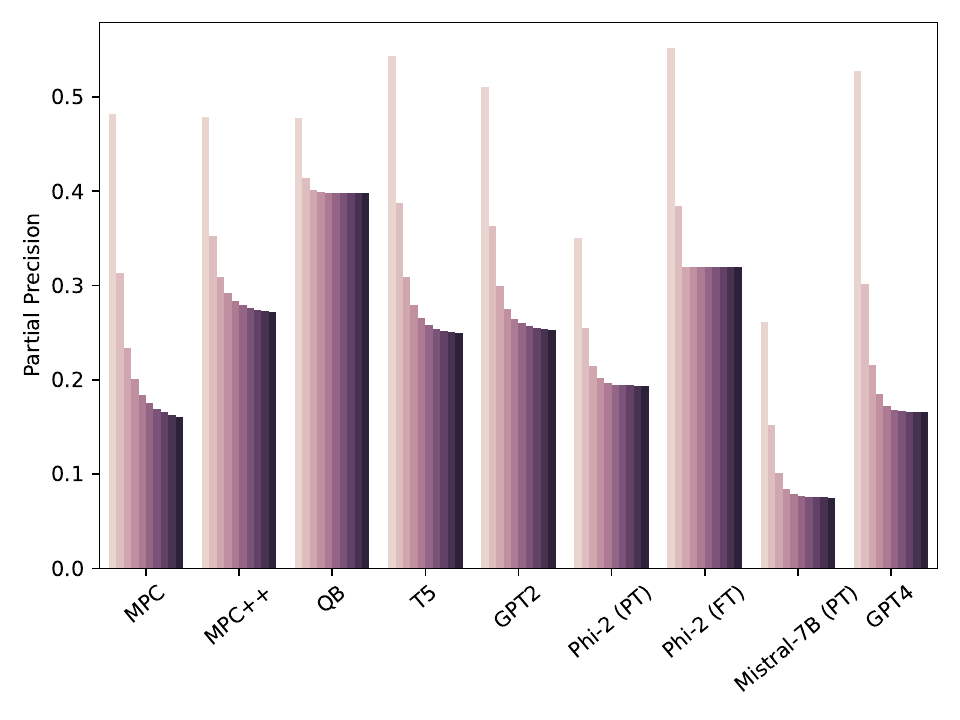}
    \caption{Truncation: \precp for DU}
    \label{fig:trunc-PrecP-DU}
\end{minipage}
\begin{minipage}{\columnwidth}
\centering
    \includegraphics[width=0.8\linewidth]{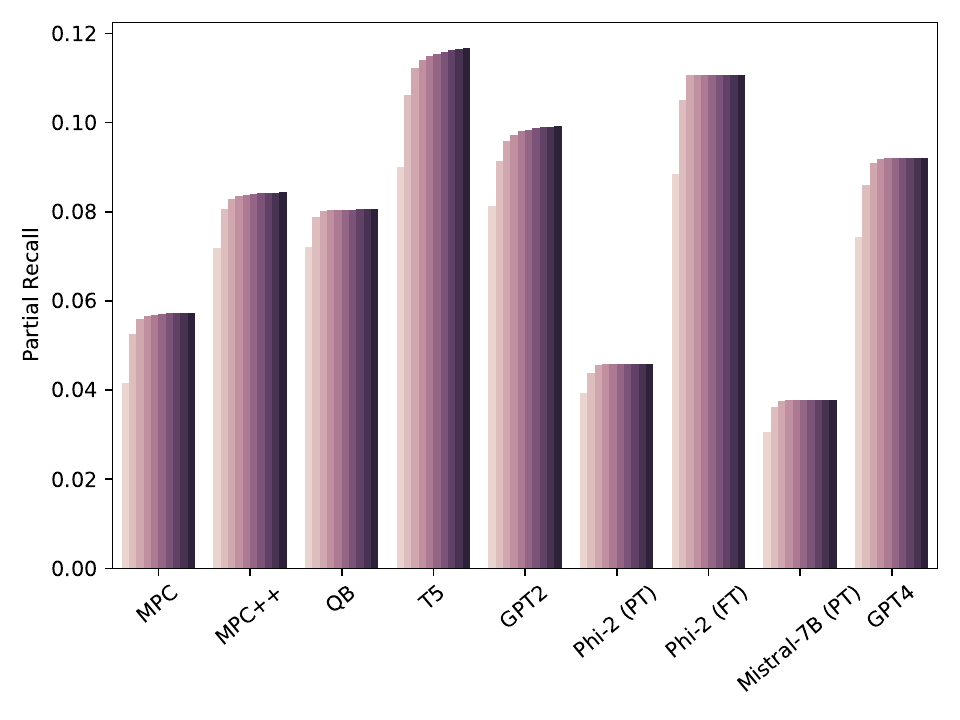}
    \caption{Truncation: \recallp for DU}
    \label{fig:trunc-RecP-DU}
\end{minipage}
\hfill
\begin{minipage}{\columnwidth}
\centering
    \includegraphics[width=0.8\linewidth]{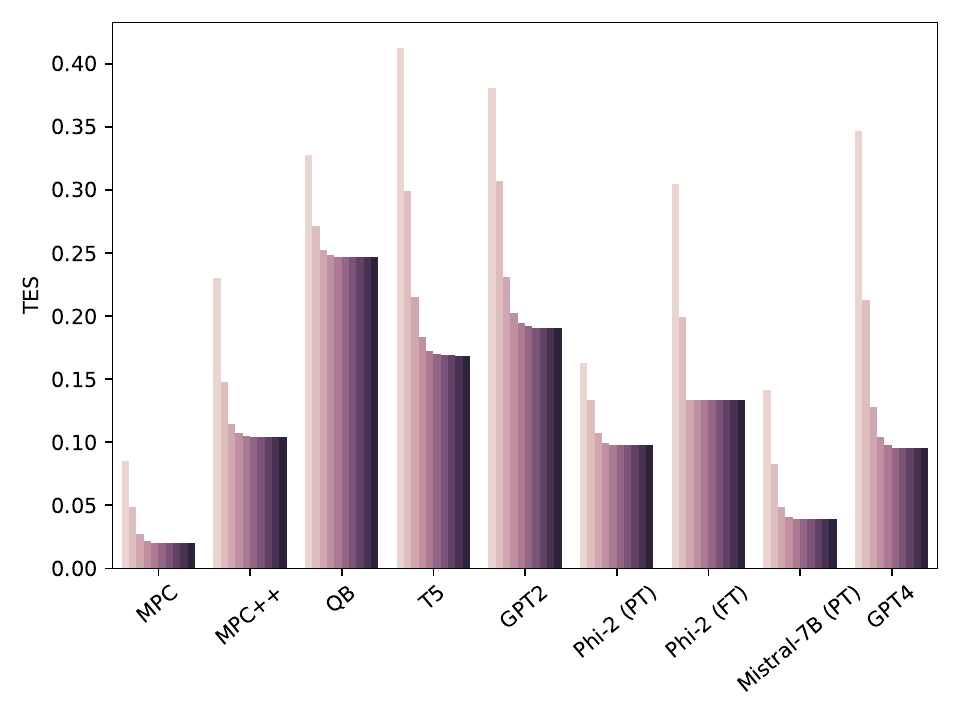}
    \caption{Truncation: TES for DU}
    \label{fig:trunc-TES-DU}
\end{minipage}
\end{figure*}

Various methods generate predictions of different sizes. Intuitively, longer suggestions have a lesser probability of being correct. That means that MR and TES should be higher for shorter suggestions. To verify this we truncate the predicted suggestions to $t$ words where we vary $t$ from 1 to 10. We show the results in Figs.~\ref{fig:trunc-MR-DD} to~\ref{fig:trunc-TES-DU} for the unseen test sets for the ``without context'' setting for both the DD and the DU datasets.

For match rate, typically there is high match rate with truncation set to 1 word and then it suddenly drops when truncation is done with 2 words. Post that the match rate continues to increase with truncation length. We believe that the match rate is high for the 1 word case because it is much easier to complete the word that the user is currently typing compared to predicting full next few words.

As expected, \precp{} and TES reduce as we increase the truncation length. Also \recallp{} increases as we increase the truncation length. Overall, these trends hold across both the datasets.


\section{Optimizing Latency in Neural Language Model-Based Ghosting}
\label{app:optimizations}

Table~\ref{tab:inferenceLatency} shows inference latency for various methods (in ms) at max TR. Note that Phi-2 finetuned latency is much smaller than the pretrained model. This is because for the pretrained models, we use 4-shot in-context learning, and generation token length is set to 40 tokens. For the Phi-2 finetuned model, we finetune using just the $\langle$context, prefix, suffix$\rangle$ without any exemplars resulting into a much shorter input. Also, max generation length for Phi-2 finetuned model is set to 5. As expected, adding context increases latency. 


We can observe that inference can be done using models like T5 and GPT2 within 100-150ms on a V100 GPU with batch size=1. While this may be within latency budget for practically deployed web scale models, further optimizations can be done to reduce latency and increase throughput. Throughput is also important when catering to millions of users with limited GPU investment. Latency with Phi-2 and Mistral models is much larger (1-3 seconds) and clearly needs to be optimized further for practical deployments.

\begin{table}[!t]
    \centering
    \scriptsize
    \begin{tabular}{|l|c|c|c|c|p{0.3\columnwidth}|}
\hline
\multirow{1}{*}{Method}&\multicolumn{1}{c|}{DD}&\multicolumn{1}{c|}{DU}&\multicolumn{1}{c|}{cDD}&\multicolumn{1}{c|}{cDU}&Hardware\\
\hline
 \textbf{QB} &3&5&80&109&\multirow{3}{0.3\columnwidth}{Intel Xeon Gold 6126 CPU}\\
 \cline{1-5}
 \textbf{\mpc} &34 &55&99&132&\\
 \cline{1-5}
 \textbf{\mpcSuff} &32 &62&105&155&\\
 \hline
 \textbf{T5 (220M)} &85&86&103&132&\multirow{5}{0.3\columnwidth}{1 V100 (32GB) with batch size=1}\\
 \cline{1-5}
 \textbf{GPT2 (124M)} &63&64&77&90&\\
 \cline{1-5}
 \textbf{Phi-2 (2.7B) (PT)} &1605 &1679&1711&1782&\\
 \cline{1-5}
 \textbf{Phi-2 (2.7B) (FT)} &367&372&478&485&\\
 \cline{1-5}
 \textbf{Mistral-7B (PT)} &2043 &2381&2315&2647&\\
 \hline
    \end{tabular}
    \caption{Inference Latency for each chat-ghosting method (in msec) at max TR. PT=Pretrained, FT=Finetuned. cDD and cDU are Contextual datasets.}
    \label{tab:inferenceLatency}
\end{table}

While our current focus was on evaluating the fundamental performance characteristics of various modeling strategies including utilization of SLMs or LLMs, we recognize the crucial importance of techniques like model pruning, distillation, and caching for practical deployment in real-time scenarios~\cite{gupta2022compression}. We could perform quantization aware training or post-training  quantization using methods like GPTQ~\cite{frantar2022gptq} and AWQ (Activation Aware Quantization)~\cite{lin2024awq} to quantize models to 4, 8 or 16 bits. We could also perform CUDA-level optimizations like ONNX optimization~\cite{jin2020compiling} (which builds a universal CUDA graph and fuses different sub-module operations), Tensor-RT LLM~\cite{tensorrt_llm} (which supports quantization  to 4/8 bit weights and 16 bit activations for Nvidia GPUs), vLLM (with Paged-attention)~\cite{kwon2023efficient}, FasterTransformer~\cite{chelba2020faster}, Deepspeed FastGen~\cite{holmes2024deepspeed}, and Llamma.cpp~\cite{llama_cpp}. Of course, pruning the models is an orthogonal direction to explore and previous research~\cite{han2015deep} has shown that pruning and quantization are complementary. Lastly, we could distill the large neural models to their smaller pretrained variants using popular knowledge distillation methods~\cite{gupta2022compression}.

For faster decoding, there are a large number of heuristics that could be leveraged like FasterTransformer~\cite{chelba2020faster}, setting a maximum prefix length, reducing beam width, and performing FP16 decoding. We could also batch requests to increase throughput. But batching could lead to slight increase in latency. A dynamic batch size could strike a good tradeoff by providing high throughput during high traffic periods and low latency during low traffic periods.

\section{Semantic Evaluation}
\label{app:semantics}
Our study mainly focuses on syntactic metrics to assess the quality of various ghosting methods. However, several predictions may not have syntactic match but they could be semantically meaningful. Hence, we conducted human evaluation of a set of 100 instances each from seen and unseen test set of the \ddc dataset. The evaluation was done for outputs from each of our 10 methods. We also obtained GPT4o evaluation for the same set. We note that GPT4o evaluation has a correlation of 0.57 with human evaluation with p-value of 2.2e-88, suggesting that GPT4o can be used as a replacement for human evaluation on the entire set. We also calculated METEOR score as a measure for semantic similarity. METEOR's effectiveness as a metric was demonstrated by the strong correlation between its ranking of suggestions and the ranking produced by GPT-4o, as measured by the length of the Longest Common Subsequence (LCS).

We used the following GPT-4o prompt for evaluation.

\begin{lstlisting}[style=prompt]
<|im_start|>system
[system](#instructions)
# ChatGhosting: Evaluating Auto-Completion Systems for Chatbots

The primary purpose of this human evaluation is to assess the quality and effectiveness of chat autocompletion models.

## Instructions

Your task is to evaluate how helpful and relevant the autocompletion predictions are by rating them on a integer scale of 1 to 5. 
Imagine you are actually typing the message and considering these suggestions.

1: Bad completion, so not helpful or relevant.
2: Some other completion, not helpful or relevant.
3: Not exactly the same intent, but still helpful or relevant.
4: Doesn't match the ground truth but means the same thing, so user can still use it.
5: Matches the ground truth exactly.

Please rate the following completions based on the above criteria.

## Format

For each candidate response, provide a rating from 1 to 5. Use json format to provide the ratings. 
Along with each rating, provide a brief explanation of why you chose that rating.

## Examples

```json
{
    "Model 1": {
        "rating": 4,
        "reason": "The completion is not exactly the same but has the same intent."
    }
    "Model 2": {
        "rating": 5,
        "reason": "The completion matches the ground truth exactly."
    }
    ...
    "Model 10": {
        "rating": 3,
        "reason": "The completion is not exactly the same intent but still helpful."
    }
}
```
<!|im_end|>
<|im_start|>user
## Input Context
Context: #context#
Ground Truth: #ground_truth#

## User Input
Prefix Typed by User: #prefix#

## Candidate AutoCompletions
Model 1: '#model_1#'
Model 3: '#model_3#'
Model 5: '#model_5#'
Model 7: '#model_7#'
Model 9: '#model_9#'
Model 11: '#model_11#'
Model 2: '#model_2#'
Model 4: '#model_4#'
Model 6: '#model_6#'
Model 8: '#model_8#'
Model 10: '#model_10#'
Model 12: '#model_12#'

<!|im_end|>
<|im_start|>assistant
[assistant](#evaluation)
```json
\end{lstlisting}

\begin{figure}
    \centering
    \includegraphics[width=\linewidth]{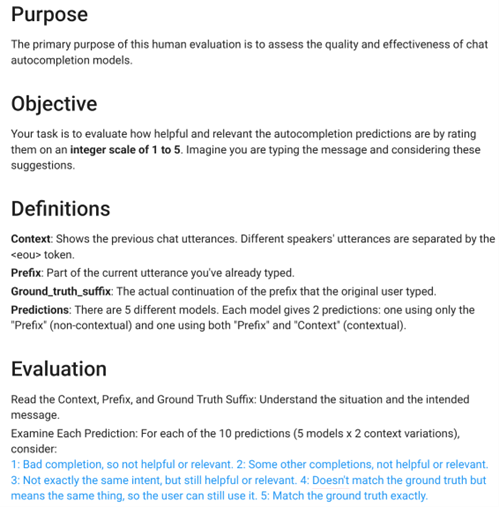}
    \caption{Human Evaluation Task Snapshot}
    \label{fig:annotationRev}
\end{figure}

Correspondingly, for human evaluation the task given to the annotators is shown in Fig.~\ref{fig:annotationRev}.

Hence, we evaluate all models on the entire test set of the \ddc dataset using both GPT4o-prompt and METEOR. We show the results in Table~\ref{tab:semanticResults}. The table shows the rankings of various methods. We observe that context helps overall from a semantic match perspective. Thus, human evaluation, GPT4o evaluation, and semantic metrics suggest that contextual models are better than non-contextual models. 

\begin{table}[!t]
    \centering
    \scriptsize
    \begin{tabular}{|l|l|l|l|l|}
    \hline
Rank&Unseen (GPT4)&Seen (GPT4)&Seen (METEOR)&Unseen (METEOR)\\
\hline
\hline
1&Context+T5&Context+MPC&Context+\mpcSuff&Context+T5\\
\hline
2&Context+GPT2&Context+\mpcSuff&Context+MPC&T5\\
\hline
3&T5&MPC&MPC&Context+GPT2\\
\hline
4&QB&\mpcSuff&\mpcSuff&GPT2\\
\hline
5&Context+QB&Context+T5&Context+T5&QB\\
\hline
6&GPT2&T5&T5&Context+\mpcSuff\\
\hline
7&Context+\mpcSuff&Context+GPT2&Context+GPT2&Context+QB\\
\hline
8&\mpcSuff&GPT2&GPT2&\mpcSuff\\
\hline
9&Context+MPC&QB&QB&Context+MPC\\
\hline
10&MPC&Context+QB&Context+QB&MPC\\
\hline
    \end{tabular}
    \caption{Semantic evaluation of our Chat-Ghosting methods on \ddc test sets.}
    \label{tab:semanticResults}
\end{table}

\section{Frequently asked questions}
\label{app:faq}
\subsection{Why did we mainly focus on prefix match based metrics?}
Standard metrics (BLEU, ROUGE, METEOR, perplexity) are not directly applicable for our task because: (1) it is important for us to match prefix (first few chars) of the suffix. Std metrics give equal importance to all words. (2) Word order is important for our task. Standard metrics ignore word order. (3) Generation of exact ground truth (and not just assigning high probability to those tokens) is important.

Semantic match based metrics like METEOR and BERTScore are becoming popular for various natural language generation tasks. However, this does not hold for Autocompletion systems. In Autocompletion, the user typically has a completion in mind, and they are splitting their cognitive bandwidth between typing and observing matching suggestions. Doing a semantic match with the shown suggestions leads to high cognitive load. Hence, if the suggestions are not syntactic exact matches, highly likely the user will not click. Hence, we mainly use syntactic prefix matching based metrics, rather than semantic matching ones.

However, for sake of completeness we also reported results using METEOR and GPT-4o evaluation in Appendix~\ref{app:semantics}.

\subsection{Why did we use DialogCC, DSTC7-Ubuntu, Open Assistant and ShareGPT datasets?}

We have evaluated our proposed methods using one task-specific dataset (DSTC7-Ubuntu) and one general open-ended conversations dataset (DialogCC). We chose DSTC7-Ubuntu because it is a standard dataset widely used in the dialog modeling literature. Also, DSTC7-Ubuntu is focused on Ubuntu technical domain, and we chose it to assess the degree to which our methods can handle technical difficulties of autocompleting code compared to other tasks. Further, DialogCC is a general open-ended conversations dataset and its purpose is to judge the performance of the model (and its generalizability) across dialogues of different domains and genres.

Although DD and DU are popular benchmark datasets in the dialog modeling community, they involve human-human conversations only. We also wanted to test our methods with human-bot conversations. Hence, we also experiment with two other datasets: Open Assistant and ShareGPT. The bot utterances are typically much larger. We have taken care to generate train and test samples from the human utterances only. Open Assistant is a human-annotated assistant conversation corpus. ShareGPT contains user-LLM-chatbots conversations collected by the ShareGPT API.

\subsection{Why is TES a useful metric?}
TES literally captures  number of keystrokes saved which is the main goal of autocompletion systems. While other metrics are at prefix level, TES is computed at utterance level. Interestingly, sometimes TES may decrease even when MR increases. Hence, a good system is one which has both high MR as well as high TES. E.g., consider the query abcde where system A generates x (some mismatching completion) as completion for prefixes a, abc, abcd and abcde. For prefix=ab, it generates cde as suffix. Here, MR=1/4, TES=3/5. Consider system B that generates x (some mismatching completion) as completion for prefixes a and ab. For prefix=abc and abcd, it generates de and e as suffix respectively. Here, MR=2/4 but TES=2/5. System A has higher TES but lower MR compared to system B.

\subsection{Why not have a metric to measure impact due to incorrect suggestions?}
Incorrect suggestions can indeed degrade the user experience, and any robust metric should ideally account for this. The TES metric implicitly penalizes incorrect suggestions. This is because, by design, TES only considers characters saved when the suggested prefix is an actual prefix of the ground truth. If a suggestion is incorrect, the user will be forced to type additional character(s) to correct or complete their input. In these cases, there would be no reduction in keystrokes, which means that incorrect suggestions result in a lower TES value, thereby providing a form of penalty. 

But incorrect suggestions could also lead to cognitive verification and suggestion-rejection load. TES does not explicitly capture the cognitive burden associated with examining suggestions. However, objectively quantifying this cognitive load is a significant challenge without access to user interaction data. The user typing characteristics, like slow typing, slow reading and examining will highly influence the cognitive load experienced. Therefore, we chose to focus on the directly measurable aspect of typing effort reduction as a more tractable starting point for our research. 

In summary, while TES provides a valuable measure of keystroke reduction and inherently penalizes incorrect suggestions, we recognize its limitations in capturing cognitive load. Our future work aims to explore methods for refining metrics to better address this challenge.

\end{document}